\documentclass[final,5p,times,twocolumn]{elsarticle}

\usepackage{amssymb}
\usepackage{amsmath}
\usepackage{latexsym}

\usepackage{url}
\usepackage{xcolor}
\usepackage{booktabs}
\usepackage{hyperref}
\usepackage{graphicx}
\usepackage{pdflscape}
\usepackage{subcaption}
\usepackage{comment}
\usepackage{float}
\usepackage{afterpage}
\usepackage{enumitem}
\usepackage{import}
\usepackage{tikz}
\usepackage{lipsum}
\usepackage[most]{tcolorbox}

\usepackage{multirow}

\journal{XXX}

\begin{document}

\begin{frontmatter}


\title{Scaling up self-supervised learning for improved surgical foundation models\tnoteref{tnote1}}%

\author[1]{Tim J.M. Jaspers\corref{cor1}\fnref{fn1}}
\cortext[cor1]{Corresponding author} 
\ead{t.j.m.jaspers@tue.nl}

\author[2]{Ronald L.P.D. de Jong\corref{cor2}\fnref{fn1}}
\author[2]{Yiping Li}
\author[1]{Carolus H.J. Kusters}
\author[5]{Franciscus H.A. Bakker}
\author[3]{Romy C. van Jaarsveld}
\author[3]{Gino M. Kuiper}
\author[3]{Richard van Hillegersberg}
\author[3]{Jelle P. Ruurda}
\author[4]{Willem M. Brinkman}
\author[2]{Josien P.W. Pluim}
\author[1]{Peter H.N. de With}
\author[2]{Marcel Breeuwer}
\author[2]{Yasmina  Al Khalil}
\author[1]{Fons van der Sommen}

\fntext[fn1]{Both authors attributed equally}

\address[1]{Department of Electrical Engineering, Video Coding \& Architectures, Eindhoven University of Technology, Eindhoven, The Netherlands}
\address[2]{Department of Biomedical Engineering, Medical Image Analysis, Eindhoven University of Technology, Eindhoven, The Netherlands}
\address[3]{Department of Surgery, University Medical Center Utrecht, Utrecht, The Netherlands}
\address[4]{Department of Oncological Urology, University Medical Center Utrecht, Utrecht, The Netherlands}
\address[5]{Department of Urology, Catharina Hospital, Eindhoven, The Netherlands}

\begin{abstract}
Foundation models have revolutionized computer vision by achieving vastly superior performance across diverse tasks through large-scale pretraining on extensive datasets. However, their application in surgical computer vision has been limited. This study addresses this gap by introducing SurgeNetXL, a novel surgical foundation model that sets a new benchmark in surgical computer vision. Trained on the largest reported surgical dataset to date, comprising over 4.7 million video frames, SurgeNetXL achieves consistent top-tier performance across six datasets spanning four surgical procedures and three tasks, including semantic segmentation, phase recognition, and critical view of safety~(CVS) classification. Compared with the best-performing surgical foundation models, SurgeNetXL shows mean improvements of 2.4, 9.0, and 12.6 percent for semantic segmentation, phase recognition, and CVS classification, respectively. Additionally, SurgeNetXL outperforms the best-performing ImageNet-based variants by 14.4, 4.0, and 1.6 percent in the respective tasks. In addition to advancing model performance, this study provides key insights into scaling pretraining datasets, extending training durations, and optimizing model architectures specifically for surgical computer vision. These findings pave the way for improved generalizability and robustness in data-scarce scenarios, offering a comprehensive framework for future research in this domain. All models and a subset of the SurgeNetXL dataset, including over 2 million video frames, are publicly available at: \url{https://github.com/TimJaspers0801/SurgeNet}.
\end{abstract}


\end{frontmatter}
\section{Introduction}
\label{sec:introduction}
In recent years, foundation models in natural image analysis have achieved groundbreaking success, reshaping the field of computer vision. These models have demonstrated remarkable capabilities across diverse tasks, including object segmentation~\citep{sam, zou2023segment}, depth estimation~\citep{depthanything}, and multi-object tracking~\citep{wang2023tracking}. By consistently achieving state-of-the-art~(SOTA) performance on a wide array of benchmarks, these models have set new standards in the field. The success of these models is largely attributed to the extensive scale at which they are trained utilizing vast datasets, cutting-edge hardware, and enormous computational resources. This approach enables them to learn rich, generalized representations that can be effectively transferred to new, unseen tasks, making these models highly versatile and impactful across various real-world applications.

Inspired by these advancements, the medical imaging community has begun to explore similar foundational approaches. Researchers are increasingly leveraging transfer learning from general-purpose models that are fine-tuned for specific medical tasks~\citep{wu2023medicalsamadapteradapting, MammoSAM}, as well as developing new models trained from scratch on large-scale medical datasets~\citep{MEDSAM, Chen2024, Pai2024, BOERS2024, Vorontsov2024}. These models have shown considerable promise in enhancing diagnostic accuracy, standardizing medical image analysis, and reducing the need for large, annotated datasets. Despite this progress, their potential in surgical applications remains largely untapped.

Current efforts in the surgical domain include a large-scale exploratory study on self-supervised learning (SSL) methods in surgical computer vision~\citep{RAMESH2023}, SSL pretraining on extensive private and public datasets for foundation models~\citep{wang2023foundation, Hirsch2023}, and investigations into the impact of dataset composition on pretraining outcomes~\citep{alapatt2023}. While promising, these efforts remain limited in scale compared to those in the natural image domain. Furthermore, there is a pressing need for comprehensive, large-scale evaluation, analysis, and benchmarking of foundation models specifically designed for surgical computer vision applications. The lack of such efforts hampers the creation of standardized benchmarks, complicates the identification of best practices, and ultimately limits the full potential of foundation models to enhance surgical outcomes.

\begin{figure}[h]
     \centering
     \includegraphics[width=0.49\textwidth]{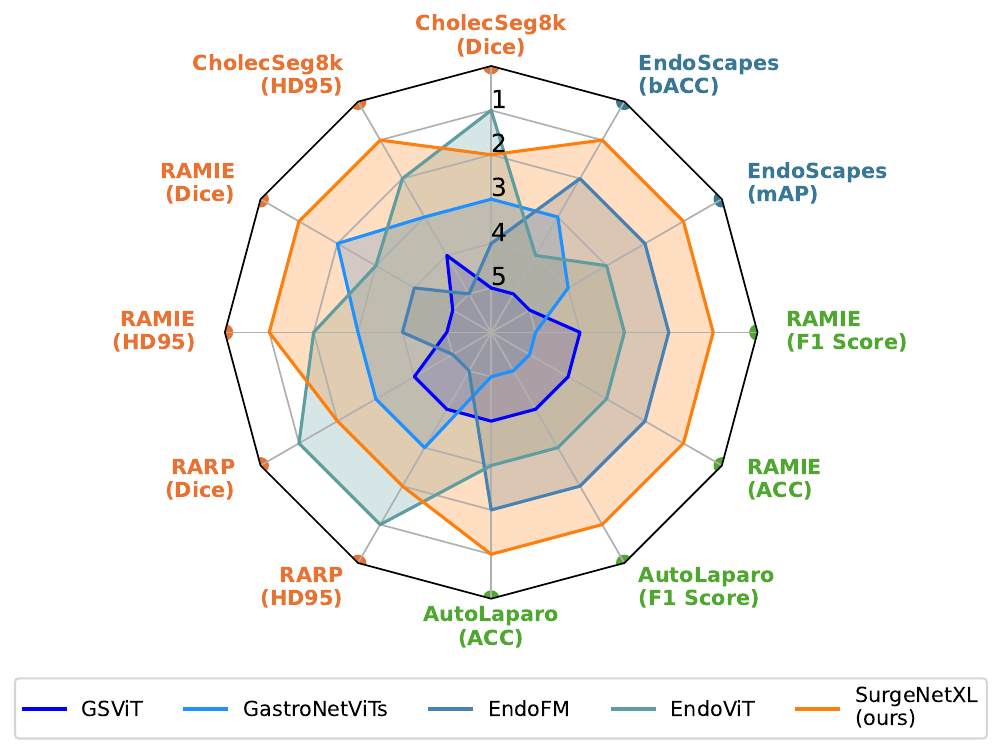}
     \caption{Radar plot showing ranks across datasets and metrics, with results from the four evaluated open-source foundation models and the proposed SurgeNetXL model. A rank of 1 indicates the best performance, while a rank of 5 indicates the worst performance. Semantic segmentation, phase recognition, and classification are shown in orange, green, and blue, respectively. This color-coding is used throughout the rest of the paper to aid clarity and consistency.}
     \label{fig:radar_rank}
\end{figure}

To address a critical gap in surgical computer vision, we establish a benchmark for foundation models and introduce a SOTA surgical foundation model that excels across diverse tasks and procedures. Fig.~\ref{fig:radar_rank} highlights the superior rankings of our model, demonstrating consistency across multiple procedures. Beyond achieving SOTA performance, this work provides valuable insights into large-scale SSL for surgical computer vision, addressing challenges in dataset diversity, training scalability, and model architecture design. Evaluated on six surgical datasets, our findings show that scaling dataset diversity, training duration, and model complexity significantly enhances performance, offering a roadmap for future advancements in this emerging field. The main contributions of this work are summarized as follows:

\begin{itemize}
\item Effectiveness of SSL for surgical computer vision demonstrated using the largest dataset reported to date.
\item Strong generalization and robust evaluation shown across six surgical datasets, four procedures, and three tasks, outperforming current SOTA foundation models.
\item Providing insights into large-scale SSL for surgical computer vision in terms of scaling, pretraining time, dataset composition, and model architecture. 
\item Release of the models and a curated dataset of 2.1M surgical video frames, establishing a critical resource for advancing surgical foundation model training.

\end{itemize}

This work significantly extends our earlier work~\citep{jaspers2024exploring} in four key aspects: (1)~adding SOTA comparisons, (2)~incorporating additional downstream tasks, (3)~providing deeper insights through comprehensive ablation studies, and (4)~substantially increasing the dataset size to further enhance the foundation model's performance. The paper is structured as follows: Section~\ref{sec:related_work} reviews related work on surgical computer vision and SSL. Section~\ref{sec:experimental_setup} outlines the experimental setup, while Section~\ref{sec:self_supervised_pretraining} provides an overview of the datasets and methods used for SSL-based pretraining. Section~\ref{sec:downstream_training} focuses on the downstream training process, and Section~\ref{sec:evaluation} explains the evaluation procedure. Section~\ref{sec:results} presents the experimental results, which are discussed further in Section~\ref{sec:discussion}. Lastly, conclusions are drawn in Section~\ref{sec:conclusion}.

\section{Related work}
\label{sec:related_work}

\subsection{Surgical computer vision}
Recent advances in surgical computer vision have focused primarily on tasks such as anatomy recognition~\citep{Mascagni2022, denBoer2023, Bakker2024}, detecting critical surgical entities such as instruments and their corresponding actions~\citep{Cholec80, Jin2018, nwoye2020}, and evaluating procedural quality metrics, such as the critical view of safety~\citep{Mascagni2022, Murali2024}. In addition, significant efforts have been made to recognize surgical phases~\citep{PADOY2012, Yitong2022, Ward2021, ban2021}, predict the remaining duration of surgeries~\citep{twinanda2019}, and leverage computer vision to enhance surgical training~\citep{HASHIMOTO2017}.

Despite their potential for minimally invasive surgery, computer vision applications in this domain have had a modest impact compared to their successes in fields such as pathology and radiology, where advanced technologies are nearing market readiness~\citep{MAIERHEIN2022}. The prevailing consensus is that surgical computer vision is still in its early stages~\citep{denBoer2022, MAIERHEIN2022}, primarily due to the scarcity of comprehensive and representative annotated datasets~\citep{MAIERHEIN2022}.

To overcome the challenge of limited availability and scarcity of annotated datasets, SSL has emerged as a promising solution. For example, \cite{RAMESH2023} conducted a large-scale benchmark study that explored the feasibility of SSL methods in the surgical domain, offering valuable insights into methodologies, training settings, and frame sampling rates. Recent research has also focused on leveraging SSL pretraining to achieve SOTA performance in surgical computer vision~\citep{wang2023foundation, Hirsch2023, Batic2024}, as well as developing SSL objectives specifically tailored to surgical applications~\citep{chen2020, SHAO2022, SurgNetTMI}. The study by~\cite{alapatt2023} is particularly relevant to our work, as it highlights the importance of dataset diversity and relevance in pretraining data composition for improving model performance.

Recent advances in surgical computer vision also include the development of vision-language models~\citep{yuan2024learningmultimodalrepresentationswatching, yuan2024procedureawaresurgicalvideolanguagepretraining}. These models integrate visual and textual information, enabling applications such as surgical report generation and multimodal understanding of procedures. While these advancements hold significant potential, their exploration lies beyond the scope of this work.

\subsection{Medical in-domain self-supervised pretraining}
SSL has emerged as a transformative approach in medical image analysis, addressing the pervasive challenge of limited annotated datasets. Traditionally, reliance on labeled data has restricted the development of generalizable deep learning models across various medical imaging modalities. By leveraging vast amounts of unlabeled data, SSL enables the development of robust models capable of excelling in diverse medical applications.

Although the viability of SSL on standard image classification datasets is a relatively recent advancement, its adoption in the medical domain is growing rapidly. Some studies focus on designing specific pretext tasks tailored to unique challenges in medical imaging~\citep{bai2019selfsupervisedlearningcardiacmr, zhuang2019selfsupervisedfeaturelearning3d, SurgNetTMI}. Others rely on proven contrastive learning techniques from general computer vision, such as DINO~\citep{caron2021emerging}, MoCo~\citep{he2020momentumcontrastunsupervisedvisual}, and SimCLR~\citep{chen2020}, to perform in-domain pretraining on medical datasets~\citep{zhou2020comparinglearnsurpassingimagenet, sowrirajan2021mococxrmocopretrainingimproves, ghesu2022selfsupervisedlearning100million}.
These efforts represent the closest body of work to our study. In this context, in-domain refers to training models on data specific to the field they are intended to perform in, which in this case means using medical images instead of natural image datasets. For instance, \cite{sowrirajan2021mococxrmocopretrainingimproves} utilized MoCo pretraining for classification tasks on the CheXpert dataset via linear evaluation, showcasing SSL's effectiveness in chest X-ray analysis. Extending this idea, \cite{ghesu2022selfsupervisedlearning100million} performed pretraining on a large-scale dataset of 100 million medical images spanning radiography, computed tomography, magnetic resonance imaging, and ultrasonography. Their results demonstrated that SSL-pretrained models not only surpassed SOTA-supervised alternatives, but also exhibited enhanced robustness to data augmentations and faster convergence during training. Similarly, \cite{BOERS2024} highlighted the critical role of in-domain data in SSL, demonstrating that pretraining on more than 5 million gastrointestinal images led to significantly better downstream performance compared to models that are pretrained on ImageNet, or its variants across various medical imaging tasks. Additionally, \cite{azizi2021bigselfsupervisedmodelsadvance} applied SSL on dermatology photos and chest X-ray scans, reporting improvements over strong baselines. Their findings showed that SSL-pretrained models were not only more robust to distribution shifts but also required fewer labeled samples to achieve competitive performance, highlighting SSL’s potential in resource-constrained settings.

\subsection{Position of our work}
Prior research on surgical foundation models has primarily compared these models to ImageNet-initialized variants \cite{Batic2024, schmidgall2024general, wang2023foundation}. These studies, published in quick succession, have yet to be directly evaluated against each other. Moreover, they have largely focused on demonstrating the benefits of in-domain pretraining over natural datasets, with limited exploration of the underlying effects. In contrast, our study expands this work by evaluating a range of publicly available foundation models—including our own—across a broad spectrum of tasks and surgical procedures, providing a more comprehensive comparison.

Building on the foundational work of \cite{RAMESH2023}, which introduced early experiments in SSL for surgical computer vision, our study takes these efforts further by conducting experiments on a significantly larger scale. While \cite{RAMESH2023} laid the groundwork with a smaller dataset, we leverage a dataset 20 times larger, providing deeper insights into the scalability and generalizability of SSL in surgical contexts. The scale of our experiments is computationally intensive but necessary to uncover the potential of SSL in surgical applications, as it enables us to analyze scalability and generalizability, paving the way for future advancements in the field.

Additionally, we expand existing open-source surgical datasets by incorporating 2.1M frames extracted from over 680 hours of surgical videos. This expansion enables SSL experiments on a scale that is comparable to those conducted on natural image datasets, allowing for a more robust evaluation. Our experiments not only provide SOTA performance, but also provide critical insights into the adaptability and scalability of SSL for surgical applications.

\section{Experimental setup} 
\label{sec:experimental_setup}

\begin{figure*}[t]
     \centering
     \includegraphics[width=0.95\textwidth]{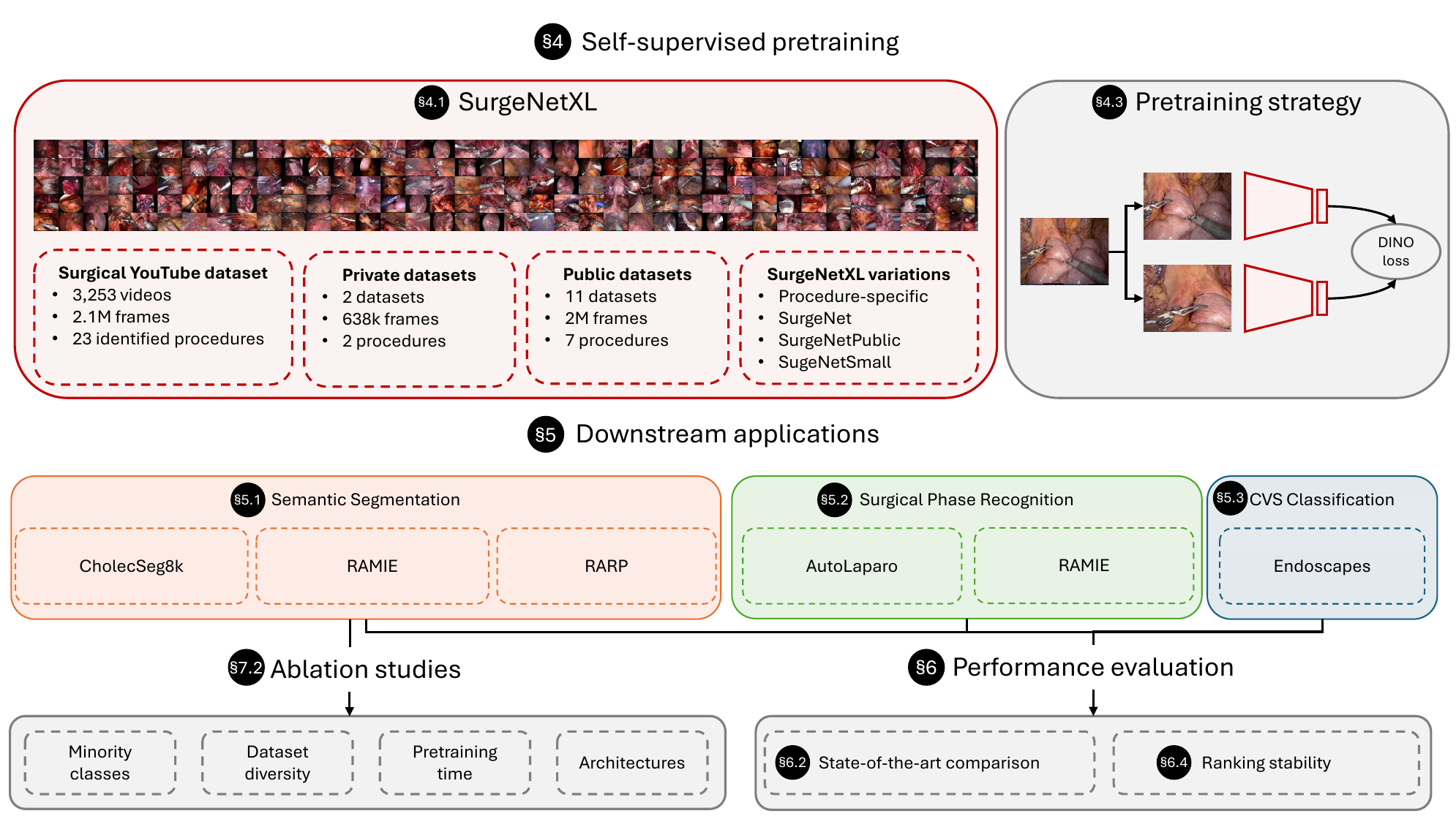}
     \caption{Overview of the experimental setup for this study. The black circles indicate the section numbers in the paper where further details about this specific aspect can be found. The datasets are color-coded for clarity, with the red section highlighting the composition of the SurgeNetXL dataset. The semantic segmentation task utilizes three datasets, phase recognition is performed using two datasets, and CVS classification is based on a single dataset. All ablation studies are focused on the semantic segmentation datasets.
}
     \label{fig:experimental_setup}
\end{figure*}

Fig.~\ref{fig:experimental_setup} illustrates the experimental framework for this study, structured into three key components. First, we perform self-supervised pretraining using the SurgeNetXL dataset, a diverse and extensive dataset described in detail in Section~\ref{sec: SurgeNetXL}, which serves as the foundation of our approach. Second, we conduct a comprehensive comparison against SOTA methods across six datasets, spanning three distinct tasks. The specifics of these tasks, datasets, and training procedures are elaborated in Sections~\ref{sec:downstream_training} and~\ref{sec:evaluation}. Finally, we conduct additional experiments on all segmentation datasets. These ablation studies, detailed in Section~\ref{sec:ablations}, offer valuable insights into the contributions of different components within our framework.

\section{Self-supervised pretraining}
\label{sec:self_supervised_pretraining}

\subsection{SurgeNetXL: a large-scale unlabeled surgical dataset}
\label{sec: SurgeNetXL}
To evaluate the impact of SSL on surgical computer vision applications, we develop SurgeNetXL, a comprehensive dataset comprising 4,711,024 frames collected from over 23 different surgical procedures. A detailed breakdown of the dataset composition is presented in Table~\ref{tab:surgenet}. SurgeNetXL integrates both public and private datasets to form a diverse and extensive collection of surgical video data. While most datasets are publicly available, additional steps are taken to enhance its size and diversity, enabling the creation of meaningful procedure-specific subsets.

\begin{table*}[t]
    \scriptsize
    \setlength{\tabcolsep}{4pt}
    \centering
    \begin{tabular}{l | l | l | c | c | c}
        \hline
        \toprule
        Procedure-specific subset           & Dataset             & Procedure  & \#videos                       & \#frames  & Public  \\
        \midrule
        \multirow{3}{*}{SurgeNetCholec}             & Cholec80 \citep{Cholec80}  &    &  76 & 179,164       & Yes       \\
                                            & HeiChole \citep{heichole}  &  Laparoscopic Cholecystectomy  &  30 & 53,427        & Yes       \\
                                            & hSDB-Chole \citep{hsdb}    &    &  24 & 18,064        & Yes       \\
        \midrule
        \multirow{1}{*}{SurgeNetRAMIE}              & RAMIE-UMCU                & RA Esophagectomy               &  28 & 377,287        & No        \\
        \midrule
        \multirow{3}{*}{SurgeNetRARP}       & ESAD  \citep{esad}         &               &  4 & 47,282        & Yes       \\
                                            & PSI-AVA \citep{psi-ava}    &    RA Prostatectomy            &  8 & 73,618        & Yes       \\
                                            & RARP-AvL                  &              &  79 & 261,516       & No        \\

        \midrule
        \multirow{6}{*}{Others}             & DSAD \citep{dsad}          & RA Rectal Resection/Extirpation & 32 & 14,623       & Yes       \\
                                            & GLENDA \citep{glenda}      & Gynecologic Laparoscopy         & 400 & 25,682        & Yes       \\
                                            & LapGyn4 (v1.2) \citep{lapgyn}  & Gynecologic Laparoscopy     & 500 & 59,616        & Yes       \\
                                            & MultiBypass140 \citep{multibypass}     & Laparoscopic Gastric bypass surgery & 140 & 749,419        & Yes       \\
                                            & hSDB-Gastric \citep{hsdb}              & RA Gastrectomy                      & 24 & 35,576        & Yes       \\
                                            & SurgToolLoc2022 \citep{surgloc}            & 11 different RA porcine procedures  & N/A & 741,516        & Yes       \\
        \midrule
        \multirow{1}{*}{Youtube}            & YouTube (\emph{ours})  & 23 identified procedures & 3,253 & 2,074,234       & Yes       \\
        \midrule
        \midrule    
        SurgeNetSmall                      & 10\% of the above (excluding YouTube)   & All of the above (excluding YouTube) & $>$1345 &  263,679   & Partly  \\
        SurgeNetPublic                      & All public datasets (excluding Youtube) & All of the above (excluding YouTube \& RA Esophagectomy) & $>$1238 & 1,997,987 & Yes     \\
        SurgeNet                            & All of the above (excluding YouTube)    & All of the above (excluding YouTube) & $>$1345 &  2,636,790 & Partly  \\
        SurgeNetXL                          & All of the above                        & All of the above                     & $>$4598 &  4,711,024 & Partly  \\
        \bottomrule
    \end{tabular}
    \vspace*{0.1cm}
    \caption{Overview of all pretraining datasets. SurgeNetXL consists of over 4.7 million frames, with the majority of the data being publicly available. SurgeNet contains more than 2.6 million frames gathered from seven distinct surgical procedures. SurgeNetPublic exclusively includes publicly available data, whereas SurgeNetSmall contains 10\% of the frames from SurgeNet, randomly selected. Additionally, the YouTube dataset, comprising over 2 million frames, is made publicly accessible. "RA" refers to robot-assisted surgery, and "partly" indicates datasets that include both public and non-public data.
}
    \label{tab:surgenet}
\end{table*}

\subsubsection{Surgical YouTube data}
Existing open-source datasets are enhanced by incorporating a curated dataset derived from surgical YouTube videos. Following the pipeline described in~\cite{schmidgall2024general}, we extract 680 hours of surgical video footage from YouTube, sampled at 1 frame per second~(fps). To ensure data quality, the curation process is conducted manually by two researchers working in the field of surgical computer vision, with non-minimally invasive procedures and out-of-body frames carefully filtered out. This results in a high-quality dataset comprising 2,074,234 frames from 23 distinct surgical procedures. To support advancements in surgical foundation models trained on large and diverse datasets, this curated dataset is now publicly available at \url{ https://github.com/TimJaspers0801/SurgeNet}.

\subsubsection{Private datasets}
SurgeNetXL is further expanded with two private datasets: one comprising robot-assisted radical prostatectomy~(RARP) procedures sourced from the Antoni van Leeuwenhoek Hospital~(AvL) in Amsterdam, and another containing robot-assisted minimally invasive esophagectomy (RAMIE) procedures from the University Medical Center Utrecht~(UMCU) in the Netherlands. The dataset consists of frames sampled from videos at 1 fps, with black borders removed to standardize the dataset. These videos are anonymized and prepared in accordance with ethical guidelines.

\subsection{SurgeNetXL variations}
From SurgeNetXL, three procedure-specific subsets are derived:

\begin{itemize}
\item \textbf{SurgeNetCholec:} this subset contains 250,655 frames extracted from laparoscopic cholecystectomy procedures.
\item \textbf{SurgeNetRAMIE:} a total of 377,287 frames are included from RAMIE procedures.
\item \textbf{SurgeNetRARP:} this subset comprises 382,416 frames from RARP procedures.
\end{itemize}

These subsets are used to analyze the impact of procedure-specific data on SSL pretraining, highlighting the importance of tailored datasets in surgical computer vision.

Additionally, for further experiments, we create three more variations of SurgeNetXL:
\begin{itemize}
    \item \textbf{SurgeNetPublic:} a subset containing only open-source datasets from SurgeNetXL. This variation excludes private datasets and the newly added YouTube-derived dataset, ensuring a purely public dataset for benchmarking.
    \item \textbf{SurgeNet:} a subset of SurgeNetXL limited to existing published datasets. SurgeNet is fully validated, and the specific procedure is known for each frame. 
    \item \textbf{SurgeNetSmall:} a randomly sampled 10\% subset of SurgeNet. Its reduced size is comparable to individual procedure-specific subsets, allowing controlled experiments to measure the trade-offs between dataset size, diversity, and performance.
\end{itemize}

\subsection{Pretraining strategy}
As SSL objective, we use the well-known framework proposed by~\cite{caron2021emerging}, “Self-Distillation with NO Labels” (DINO). This method employs distillation-based techniques to enable efficient learning with smaller batch sizes, reducing the need for extensive computational resources. Given its proven success in both general and medical imaging tasks, we focus on DINO as the pretraining strategy for this study. While alternative pretraining methods, such as SimCLR, MoCoV2, or MAE, have been explored in previous research and have shown similar benefits~\citep{BOERS2024, RAMESH2023}, we do not investigate them further due to the strong performance of DINO in the natural image domain and its demonstrated effectiveness in medical imaging. Moreover, the computational demands of experimenting with multiple strategies are substantial, so we concentrate on DINO to maintain focus and optimize our resources. Our implementation closely follows the original framework, with the exact details provided in Table~\ref{tab: dino details} in the supplementary materials.

We initiate pretraining from ImageNet-initialized weights as recommended by~\cite{RAMESH2023}, and align with the concept of fine-tuning natural computer vision foundation models for medical applications. We perform pretraining on four 40-GB A100 GPUs (NVIDIA Corp., CA, USA) using a maximum feasible batch size of~544. Due to the number of pretraining experiments, and the vast amount of data, we keep the training epochs limited to a maximum of 50~epochs.

\subsection{Model architectures}
To demonstrate the effectiveness of SSL for representation learning on surgical data, we utilize three distinct SOTA backbones: ConvNeXt~\citep{liu2022convnet}, PVTv2~\citep{wang2021pyramid}, and CAFormer~\citep{MetaFormer}, a CNN-based, transformer-based, and hybrid architecture, respectively. These models are selected to illustrate the robust applicability of SSL across fundamentally different architectural paradigms and to identify which type of architectures are best suited for SSL.

The specific variants of these models are PVTv2-B2, ConvNeXtV2-tiny, and CAFormer-S18. These versions are designed with a relatively small number of parameters, enabling faster inference times—a critical feature for many surgical applications that require real-time performance.

\section{Downstream network training}
\label{sec:downstream_training}
\begin{figure*}[h]
     \centering
     \includegraphics[width=0.71\textwidth]{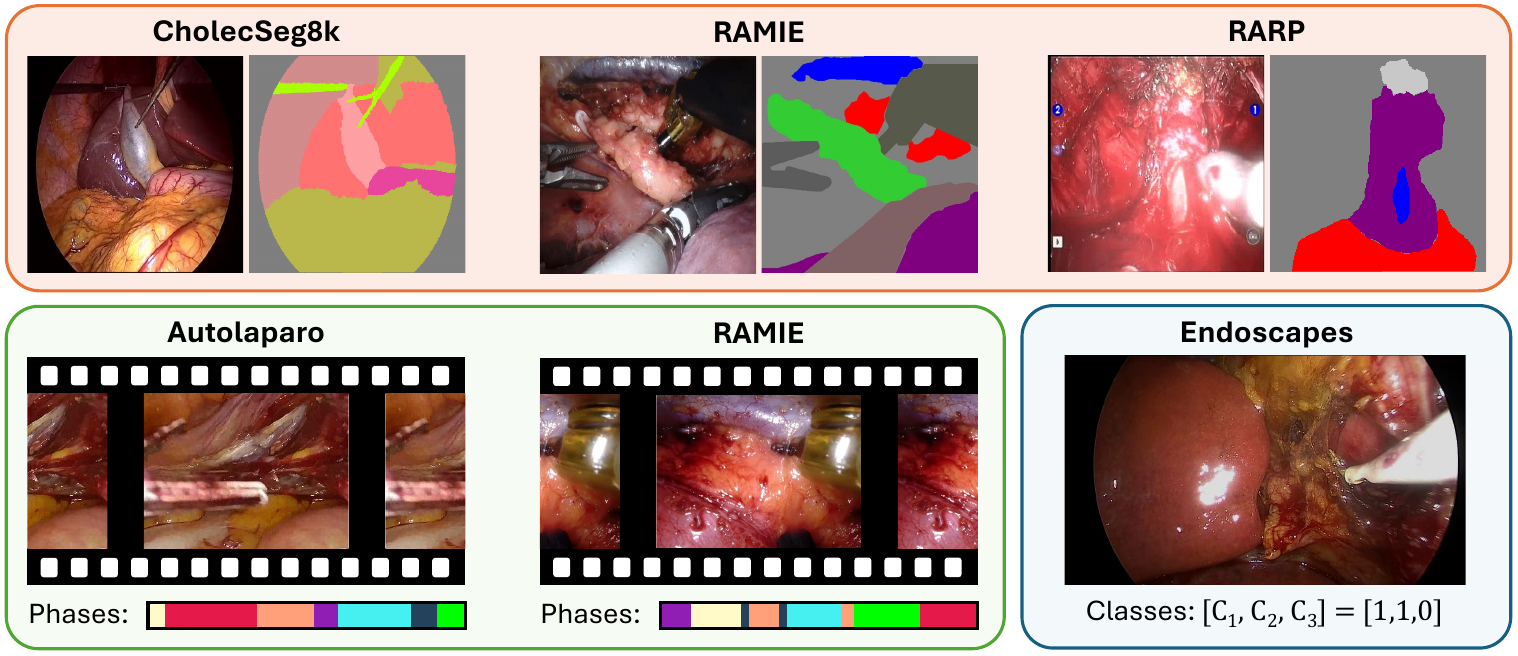}
     \caption{Visual overview of all downstream datasets, including semantic segmentation, phase recognition, and classification in orange, green, and blue, respectively.}
     \label{fig:overview_downstream_datasets}
\end{figure*}

The experiments in this study are conducted on six downstream datasets across three tasks: semantic segmentation, phase recognition, and classification. These datasets cover various types of surgeries, including cholecystectomy, prostatectomy, esophagectomy, and hysterectomy, providing a broad range of surgical scenarios. Table~\ref{tab:downstream} presents an overview of these datasets and their composition. There is no patient overlap between the pretraining datasets and the test sets in any of these downstream datasets. Fig.~\ref{fig:overview_downstream_datasets} provides a visual overview of all downstream datasets. The following sections explain the specific datasets and implementation strategies used for each task.

\begin{table*}[h]
    \scriptsize
    \setlength{\tabcolsep}{4pt}
    \centering
    \begin{tabular}{l | l | c  c | c c | c | c}
        \hline
        \toprule
        \multirow{2}{*}{Downstream Dataset} & \multirow{2}{*}{Task}          & Training  & Training  & Test     & Test     & Number of & \multirow{2}{*}{Public} \\
                                            &                                 & patients  & frames    & patients & frames  &  classes \\ 
        \midrule
        CholecSeg8k \citep{Cholecseg8k}     &  \multirow{3}{*}{Semantic segmentation}     & 13        &    6,800  &  4       &  1,280  & 8          & Yes \\
        RAMIE                              &        & 27        &    749    &  4       & 120     & 12         & No \\
        RARP                               &        & 148       &    475    &  34      &  60     & 4          & No \\
        \midrule
        AutoLaparo \citep{wang2022autolaparonewdatasetintegrated}       &  \multirow{2}{*}{Phase recognition}   & 10        &   40211     &  7       &  28060    & 7          & Yes \\
        RAMIE                              &     & 18        &    132636    &  9       & 66596     & 13         & No \\
        \midrule
        Endoscapes \citep{murali2024endoscapesdatasetsurgicalscene}                         & CVS Classification   & 161        &    9,291    &  40       & 1,799     & 3         & Yes \\
        
        \bottomrule
    \end{tabular}
    \vspace*{0.1cm}
    \caption{Descriptions of the downstream datasets. The datasets exhibit diversity in terms of patient inclusion, annotated frames, and the number of structures, offering a representative sample of surgical downstream datasets that could benefit from SSL.}
    \label{tab:downstream}
\end{table*}

\subsection{Semantic segmentation}
\subsubsection{Datasets}
Three downstream semantic segmentation datasets are used, each corresponding to a specific surgical procedure: cholecystectomy, RARP, and RAMIE. The datasets primarily focus on anatomy segmentation but also include labels for surgical tools. They are selected to represent diverse surgical scenarios, varying in data size, patient demographics, and anatomical structures. This ensures a thorough evaluation of model performance across different conditions and demonstrates the robustness and applicability of SSL to various surgical procedures.

\textbf{CholecSeg8k:} The publicly available CholecSeg8k dataset~\citep{Cholecseg8k} includes 8,080 frames from laparoscopic cholecystectomy procedures, annotated with semantic segmentation masks. Of these, we use 6,800 for training and 1,280 for testing. We exclude low-prevalence classes (\textit{blood, cystic duct, hepatic vein, and liver ligament}) to ensure a robust analysis and to stay consistent with previous studies~\citep{grammatikopoulou2024spatio, zhang2024towards}. While all patients in the CholecSeg8k training set appear in the SurgeNet pretraining dataset, this overlap does not affect downstream evaluation, as no overlap exists between the pretraining data and the CholecSeg8k test set. 

\textbf{RAMIE:} The RAMIE dataset consists of 869 labeled frames from 31 distinct patients undergoing thoracoscopic RAMIE, with 749 frames allocated for training and 120 frames reserved for testing. Further details about this dataset are explained in \cite{dejong2024benchmarking}. In this dataset, 22 of the 27 training patients are included in the SurgeNet pretraining dataset, and there is no overlap with the test set.

\textbf{RARP:} The RARP dataset includes 535 labeled frames from 208 patients, acquired from two medical centers. The training set consists of 475 frames from 148 patients, while the test set contains 60 frames from 34 patients. This dataset is an extension of the dataset used in a previous publication~\citep{Bakker2024}. Further details can be found in that publication. Note that there is no overlap between the patients in the RARP downstream dataset and the RARP videos in the SurgeNet pretraining dataset.

\subsubsection{Training details}
For segmentation tasks, the PVTv2, ConvNeXt, and CAFormer backbones are paired with a feature pyramid network~(FPN) decoder~\citep{FPN}. We select this decoder for its lightweight design, keeping the model compact and highlighting the pretrained backbone. All segmentation datasets are trained using five-fold cross-validation at the patient level, with approximately 80\% reserved for training and 20\% for validation. Training parameters are identical across all experiments. Frames are resized to 256$\times$256~pixels using bicubic interpolation. The models are trained using cross-entropy loss and the AdamW optimizer with a learning rate of 1$\times$10$^{-5}$, which is halved after 10~epochs without improvement in validation loss. All models are trained on a GeForce RTX 2080 Ti GPU (NVIDIA Corp., CA, USA) with a batch size of 16 and an early stopping criterion of 15 epochs, determined based on convergence seen in the initial experiments. Data augmentation is restricted to horizontal and vertical flipping, as well as rotation, each applied with a 50\% probability. This approach prevents overfitting to augmentation-specific features and ensures that the model focuses on adapting the pre-trained representations to the downstream task, avoiding unnecessary noise or distortion.

\subsection{Surgical phase recognition}
\subsubsection{Datasets}
\textbf{AutoLaparo:} The publicly available AutoLaparo dataset~\citep{wang2022autolaparonewdatasetintegrated} is a widely used benchmark in surgical phase recognition. It consists of full-length videos of entire hysterectomy procedures, including annotations for seven distinct phases. The dataset is divided into training (10 videos, 40,211 frames), validation (4 videos, 12,056 frames), and testing (9 videos, 12,056 frames).

\textbf{RAMIE:} The RAMIE surgical phase recognition dataset includes 27 thoracoscopic RAMIE recordings with 13 distinct phases. The dataset is divided into 14 videos for training, 4 for validation, and 9 for testing. A three-fold cross-validation strategy is applied, with patient-level distinctions defining the folds. The annotated dataset contains 132,636 frames for training and validation and 66,596 frames for testing. Further details are provided in a previous publication~\citep{li2024benchmarking}.

\subsubsection{Training details}
For the surgical phase recognition task, we adopt the two-step training strategy proposed by TeCNO~\citep{czempiel2020tecno}. In the first step, a backbone model is trained to predict surgical phases from individual frames. The second step uses a Multi-Stage Temporal Convolutional Network (MS-TCN) to refine the extracted features by incorporating temporal context. The videos are first split into frames and resized to 256$\times$256 pixels. All models are trained on a GeForce RTX 2080 Ti GPU (NVIDIA Corp., CA, USA), with identical training parameters in all experiments. In the first stage, we train a CAFormer feature extractor on individual frames using a learning rate of 1$\times$10$^{-5}$ and cross-entropy loss. In the second stage, MS-TCN is trained on the extracted features with a learning rate of 7$\times$10$^{-4}$ for 200 epochs, also using cross-entropy loss, following the implementation of \citep{RIVOIR2024103126}.

\subsection{CVS classification}
\subsubsection{Datasets}

\textbf{Endoscapes:}
The third task in this study is the classification of the Critical View of Safety (CVS), a key safety protocol in laparoscopic cholecystectomy. For this, we use the Endoscapes dataset~\citep{murali2024endoscapesdatasetsurgicalscene}, focusing on the Endoscapes-CVS201 subset of 11,090 frames annotated by three experts with binary labels for CVS achievement. The dataset includes 201 videos split into 120 training, 41 validation, and 40 testing videos, corresponding to 6,960 training, 2,331 validation, and 1,799 testing frames. We combine training and validation sets for five-fold cross-validation, ensuring patient-level splits to prevent data leakage.

\subsubsection{Training details}
We closely follow the training protocol outlined in~\cite{murali2024endoscapesdatasetsurgicalscene} to ensure consistency and comparability of results. All models are trained using five-fold cross-validation, with splits made at the patient level to prevent data leakage. Input frames are resized to a resolution of 224$\times$399 pixels using bicubic interpolation. Binary cross-entropy is used as the loss function and model parameters are optimized with a learning rate of 1$\times$10$^{-5}$, reduced by half after 10 epochs without improvement in validation loss. Training continues until convergence, with early stopping applied after 15 epochs. To improve generalization, we apply data augmentation using RandAugment~\citep{cubuk2019randaugmentpracticalautomateddata}. All experiments are conducted on a GeForce RTX 3090 Ti GPU (NVIDIA Corp., CA, USA), using a batch size of~64. These settings ensure efficient training while maintaining compatibility with prior work.

\section{Performance evaluation}
\label{sec:evaluation}
\subsection{SOTA comparison}
We compare our models to several publicly available foundation models in the domains of laparoscopy and endoscopy, which serve as benchmarks for evaluating the SurgeNetXL performance. 

\textbf{GastroNet}~\citep{BOERS2024}: Trained on over 5 million images from the gastrointestinal tract using the DINO method~\citep{caron2021emerging}. GastroNet includes two versions: a ResNet50 encoder and a ViT-small encoder~\citep{dosovitskiy2021imageworth16x16words}. Although not trained on surgical data, the images are assumed to be more closely related than natural images.
    
\textbf{EndoFM}~\citep{wang2023foundation}: Trained using DINO on over 33,000 endoscopic video clips, including colonoscopy, gastroscopy, and laparoscopy data.
    
\textbf{GSViT}~\citep{schmidgall2024general}: Trained on over 680 hours of YouTube surgical videos, utilizing all extracted frames. It adopts the EfficientViT-M5 encoder~\cite{liu2023efficientvit} and a modified version of the MAE pretraining method.

\textbf{EndoViT}~\citep{Batic2024}: Trained on over 700,000 surgical video frames using a ViT-base encoder pretrained using the Masked Autoencoder (MAE) method~\citep{he2021maskedautoencodersscalablevision}.

The exact implementation details for each of the SOTA models can be found in the supplementary materials.

\subsection{SurgeNetXL variations}
Alongside these SOTA models, we also evaluate various configurations of the proposed CAFormer architecture to investigate the influence of dataset composition and domain-specific pretraining on the model's generalizability and performance.

\textbf{In-domain Pretraining:} We compare CAFormer models pretrained on ImageNet1k and ImageNet21k to assess the impact of in-domain pretraining.

\textbf{Procedure-Specific Data:} CAFormer pretrained exclusively on laparoscopic cholecystectomy data is evaluated to understand its performance when limited to a single procedure.

\textbf{Public Surgical Data:} We assess CAFormer pretrained on all publicly available datasets in SurgeNetXL, excluding our YouTube extension, to measure the contribution of the YouTube dataset to overall performance.

\subsection{Metrics}
The performance of the models is evaluated on semantic segmentation, surgical phase recognition, and CVS classification tasks using the metrics most widely adopted for each. For segmentation, both an overlap-based metric (Dice score) and an edge-based metric (95th percentile Hausdorff distance, HD95, expressed in pixels) are employed to ensure a comprehensive evaluation. Surgical phase recognition is assessed using accuracy~(ACC) and F1 score, capturing both overall correctness and the balance between precision and recall. For CVS classification, mean average precision~(mAP) and balanced accuracy~(bACC) are used to evaluate class-wise performance and address class imbalance. This evaluation framework provides a robust analysis of model performance across diverse tasks.

\subsection{Ranking stability}
To assess the stability of model rankings across all three tasks in the context of sampling variability, we employ a rigorous bootstrapping approach as proposed by~\cite{Wiesenfarth2021}. This evaluation is particularly crucial when comparing architectures tested on relatively small datasets or a limited number of patients per task~\citep{Wiesenfarth2021, Varoquaux2022}.

As noted previously, each model is trained using a five-fold cross-validation. All predictions of the five models are used to perform the bootstrapping. We generate 1,000 bootstrap samples with replacement, ensuring the random seed governing image selection remained similar across experiments.

For each bootstrap iteration, the models are ranked from 1 to 10 based on their performance, resulting in a distribution of 1,000 ranks per model. These distributions provide a reliable way to evaluate how consistently models rank across different data samples, offering a robust measure of ranking stability despite variations caused by sampling. To visualize ranking stability, we use blob plots, where models are arranged from left (best-performing) to right (worst-performing). The y-axis represents the ranking, with lower ranks indicating better performance. These visualizations offer an intuitive way to compare the robustness of model rankings across different tasks.

\section{Results}
\label{sec:results}

\subsection{SOTA-comparison}
Table~\ref{tab:sota_results} compares SurgeNetXL with other publicly available foundation models, as well as different variants of SurgeNet. SurgeNetXL achieves top-2 performance across all evaluated datasets and metrics, demonstrating its robustness and generalizability. In contrast, other foundation models exhibit greater variability in performance. For instance, while EndoViT performs well on the CholecSeg8k dataset, its median performance on the RAMIE dataset is approximately 10\% lower in both Dice and HD95 metrics compared to SurgeNetXL. This pattern of inconsistent performance is evident across other foundation models, further emphasizing the reliability of SurgeNetXL. Moreover, while EndoViT is close in performance to our SurgeNet models on some datasets, its model size is about 3.5 times as large. This poses additional downsides, such as slower inference and training times.

In comparison with the best-performing ImageNet-based variant, SurgeNetXL demonstrates significant improvements. On semantic segmentation datasets, it achieves median Dice gains of 7.7\%, 9.5\%, and 4.8\%, respectively. For phase recognition tasks, SurgeNetXL achieves accuracy improvements of 2.9\% and 2.7\%. In CVS classification, although the mAP remains unchanged, the bACC increases by 3.2\%. These findings highlight the effectiveness of the model across diverse tasks.

Compared to its variants, SurgeNetXL demonstrates significant advantages. It consistently outperforms SurgeNetCholec across all tasks and metrics, even on the laparoscopic cholecystectomy datasets, with the exception of bACC on the Endoscapes dataset (0.63 vs. 0.64). This suggests that models tailored for specific procedures benefit from pretraining on larger and more diverse datasets. Furthermore, SurgeNetXL surpasses SurgeNetPublic in all evaluated tasks and metrics, highlighting the impact of one of our key contributions: the acquisition, curation and release of the Surgical YouTube dataset as an extension to existing datasets. This new dataset introduces greater diversity in the range of included procedures and video quality.
\begin{table*}[]
    \tiny
    \setlength{\tabcolsep}{4pt}
    \centering
    \caption{Qualitative evaluation of various SurgeNet models compared to SOTA methods on a variety of downstream datasets. Results are reported as median (min-max), resulting from cross-validation folds, with the best values indicated in bold. The code names are later used in Fig.~\ref{fig:main} to denote the various models.}
    \label{tab:sota_results}
    \begin{tabular}{l | l | c | c | l l l l l l | l l l l | l l }
        \hline
        \toprule
        Code & Method          & Backbone & Number of  &\multicolumn{2}{l}{CholecSeg8k} & \multicolumn{2}{l}{RAMIE} & \multicolumn{2}{l|}{RARP} & \multicolumn{2}{l}{AutoLaparo} & \multicolumn{2}{l|}{RAMIE} & \multicolumn{2}{l}{EndoScapes} \\ 
        name &                &          & parameters          & Dice & HD95  & Dice     & HD95         & Dice     & HD95  & ACC    & F1 score  & ACC    & F1 score      & mAP    & bACC       \\ 
        \midrule
        C1 & GastroNet       &  ResNet50    &    23.5M   & 0.65       & 55.15 & 0.60 & 54.91 & 0.63  & 23.77 & 0.69  & 0.58  & 0.72  & 0.64  & 0.44  & 0.60  \\
           &             &              &       & \textcolor{gray}{(0.65-0.66)} &  \textcolor{gray}{(52.49-58.63)} & \textcolor{gray}{(0.57-0.61)} & \textcolor{gray}{(51.84-62.29)} & \textcolor{gray}{(0.62-0.68)} & \textcolor{gray}{(21.79-25.92)}  &\textcolor{gray}{(0.69-0.70)}  &\textcolor{gray}{(0.58-0.59)}  &\textcolor{gray}{(0.68-0.75)}  &\textcolor{gray}{(0.59-0.68)}  & \textcolor{gray}{(0.41-0.47)} & \textcolor{gray}{(0.59-0.65)}  \\
        C2 & GastroNet       &  ViT-Small   &    21.1M   & 0.63 & 55.71 & 0.66 & 45.53 & 0.64 & 22.06 & 0.61  & 0.54  & 0.70  & 0.61  & 0.40 & 0.59 \\
           &             &              &       & \textcolor{gray}{(0.59-0.67)} & \textcolor{gray}{(51.75-56.92)} & \textcolor{gray}{(0.64-0.67)} & \textcolor{gray}{(42.64-47.19)} & \textcolor{gray}{(0.62-0.68)} & \textcolor{gray}{(20.14-23.82)} &\textcolor{gray}{(0.61-0.62)}  &\textcolor{gray}{(0.54-0.54)}  &\textcolor{gray}{(0.70-71)}  &\textcolor{gray}{(0.60-0.63)}  & \textcolor{gray}{(0.38-0.44)} & \textcolor{gray}{(0.56-0.60)}   \\
        C3 & Endo-FM         &  ViT-Base   &    85.8M    & 0.54 & 76.32 & 0.51 & 77.65 & 0.45 & 62.47 & 0.79  & 0.67  & 0.80  & 0.72  & 0.45 & 0.59  \\
           &             &              &       & \textcolor{gray}{(0.52-0.60)} & \textcolor{gray}{(75.46-81.92)} & \textcolor{gray}{(0.44-0.53)} & \textcolor{gray}{(73.73-87.54)} & \textcolor{gray}{(0.42-0.48)} & \textcolor{gray}{(49.22-76.51)} &\textcolor{gray}{(0.78-0.79)}  &\textcolor{gray}{(0.66-0.68)}  &\textcolor{gray}{(0.80-0.81)}  &\textcolor{gray}{(0.71-0.73)}  & \textcolor{gray}{(0.41-0.50)}  & \textcolor{gray}{(0.53-0.63)}  \\
        C4 & GSViT           &  EfficientViT-M5  & 12.1M  & 0.53 & 70.41 & 0.49 & 71.29 & 0.45 & 39.44 & 0.69  & 0.60  & 0.74  & 0.64  & 0.22 & 0.51 \\
           &             &                   &   & \textcolor{gray}{(0.50-0.57)} & \textcolor{gray}{(56.78-79.72)} & \textcolor{gray}{(0.48-0.51)} & \textcolor{gray}{(65.40-79.35)} & \textcolor{gray}{(0.45-0.50)} & \textcolor{gray}{(33.85-42.52)} &\textcolor{gray}{(0.67-0.71)}  &\textcolor{gray}{(0.59-0.62)}  &\textcolor{gray}{(0.73-0.75)	}  &\textcolor{gray}{(0.64-0.65)}  & \textcolor{gray}{(0.20-0.26)} & \textcolor{gray}{(0.46-0.55)}   \\
        C5 & EndoViT         & ViT-Base &  85.8M   &  \textbf{0.73} & 42.48 & 0.63 & 45.95 & \textbf{0.69} & \textbf{20.87} & 0.79  & 0.66  & 0.79 & 0.67 & 0.41 & 0.57  \\
           &             &    &      &\textcolor{gray}{(0.66-0.73)} & \textcolor{gray}{(39.84-49.21)} & \textcolor{gray}{(0.63-0.66)} & \textcolor{gray}{(40.23-50.53)} & \textcolor{gray}{(0.64-0.73)} & \textcolor{gray}{(20.08-27.80)} &\textcolor{gray}{(0.79-0.79)}  &\textcolor{gray}{(0.66-0.66)}  &\textcolor{gray}{(0.78-0.79)	}  &\textcolor{gray}{(0.65-0.68)} & \textcolor{gray}{(0.37-0.48)} & \textcolor{gray}{(0.55-0.60)}   \\
        \midrule
        C6 & ImageNet1k      &   &   & 0.62 & 54.90 & 0.62 & 51.59 & 0.61 & 26.64 & 0.78  & 0.70  & 0.79 & 0.71 & 0.44 & 0.62 \\
           &             &                &     & \textcolor{gray}{(0.55-0.65)} & \textcolor{gray}{(50.51-62.11)} & \textcolor{gray}{(0.61-0.63)} & \textcolor{gray}{(45.16-59.86)} & \textcolor{gray}{(0.58-0.63)} & \textcolor{gray}{(22.92-29.31)} &\textcolor{gray}{(0.77-0.79)}  &\textcolor{gray}{(0.69-0.72)}  &\textcolor{gray}{(0.79-0.80)}  &\textcolor{gray}{(0.69-0.72)}  &  \textcolor{gray}{(0.41-0.50)} &  \textcolor{gray}{(0.60-0.64)} \\
        C7 & ImageNet21k     &    &   & 0.65 & 51.45 & 0.63 & 56.63 & 0.63 & 25.31 & 0.83  & 0.70   & 0.80  & 0.72  & \textbf{0.47} & 0.61  \\
           &             &               &    & \textcolor{gray}{(0.55-0.68)} & \textcolor{gray}{(45.60-60.03)} & \textcolor{gray}{(0.62-0.64)} & \textcolor{gray}{(47.88-59.93)} & \textcolor{gray}{(0.60-0.66)} & \textcolor{gray}{(22.99-28.49)} &\textcolor{gray}{(0.83-0.83)}  & \textcolor{gray}{(0.69-0.70)}  &\textcolor{gray}{(0.80-0.80)} &\textcolor{gray}{(0.72-0.72)}  & \textcolor{gray}{(0.44-0.53)} & \textcolor{gray}{(0.58-0.68)} \\
        S1 & SurgeNetCholec \emph{(ours)} & CaFormerS18   &   24.3M   & 0.66 & 47.28 & 0.60 & 55.00 & 0.64 & 25.56 & 0.82  & 0.70 & 0.80  & 0.70  & 0.43 & \textbf{0.64}  \\
           &             &               &      & \textcolor{gray}{(0.56-0.66)} & \textcolor{gray}{(43.68-51.25)} & \textcolor{gray}{(0.58-0.61)} & \textcolor{gray}{(49.78-59.18)} & \textcolor{gray}{(0.59-0.64)} & \textcolor{gray}{(23.31-25.69)} &\textcolor{gray}{(0.82-0.83)}  &\textcolor{gray}{(0.70-0.70)}  &\textcolor{gray}{(0.79-0.80)}  &\textcolor{gray}{(0.68-0.70)}  & \textcolor{gray}{(0.41-0.47)} & \textcolor{gray}{(0.59-0.66)}  \\
        S2 & SurgeNetPublic \emph{(ours)} &      &        & 0.67 & 44.60 & 0.67 & 42.85 & 0.65 &  23.65  & 0.83  & 0.71  & 0.80  & 0.71  & 0.46 & 0.60 \\
           &            &             &        & \textcolor{gray}{(0.63-0.71)} & \textcolor{gray}{(42.14-57.83)} & \textcolor{gray}{(0.64-0.67)} & \textcolor{gray}{(38.46-46.97)} & \textcolor{gray}{(0.62-0.69)} & \textcolor{gray}{(21.86-24.53)} &\textcolor{gray}{(0.81-0.84)}  &\textcolor{gray}{(0.68-0.73)}  &\textcolor{gray}{(0.79-0.80)}  &\textcolor{gray}{(0.70-0.72)}  & \textcolor{gray}{(0.44-0.49)} & \textcolor{gray}{(0.58-0.62)}  \\
        S3 & SurgeNetXL \emph{(ours)} &  &   & 0.70 & \textbf{40.99} & \textbf{0.69} & \textbf{37.08} & 0.66 & 21.81 & \textbf{0.85}  & \textbf{0.74}  & \textbf{0.82} & \textbf{0.75}  & \textbf{0.47} & 0.63  \\ &
                        &             &        & \textcolor{gray}{(0.64-0.73)} & \textcolor{gray}{(37.89-45.79)} & \textcolor{gray}{(0.69-0.70)} & \textcolor{gray}{(34.97-39.42)} & \textcolor{gray}{(0.63-0.68)} & \textcolor{gray}{(20.10-23.99)} &\textcolor{gray}{(0.85-0.86)}  &\textcolor{gray}{(0.72-0.75)}  &\textcolor{gray}{(0.82-0.83)} &\textcolor{gray}{(0.73-0.76)}  & \textcolor{gray}{(0.45-0.49)} & \textcolor{gray}{(0.60-0.65)}  \\
        \bottomrule
    \end{tabular}
\end{table*}

\subsubsection{Ranking stability}
\begin{figure}[t]
     \centering
     \includegraphics[width=0.5\textwidth]{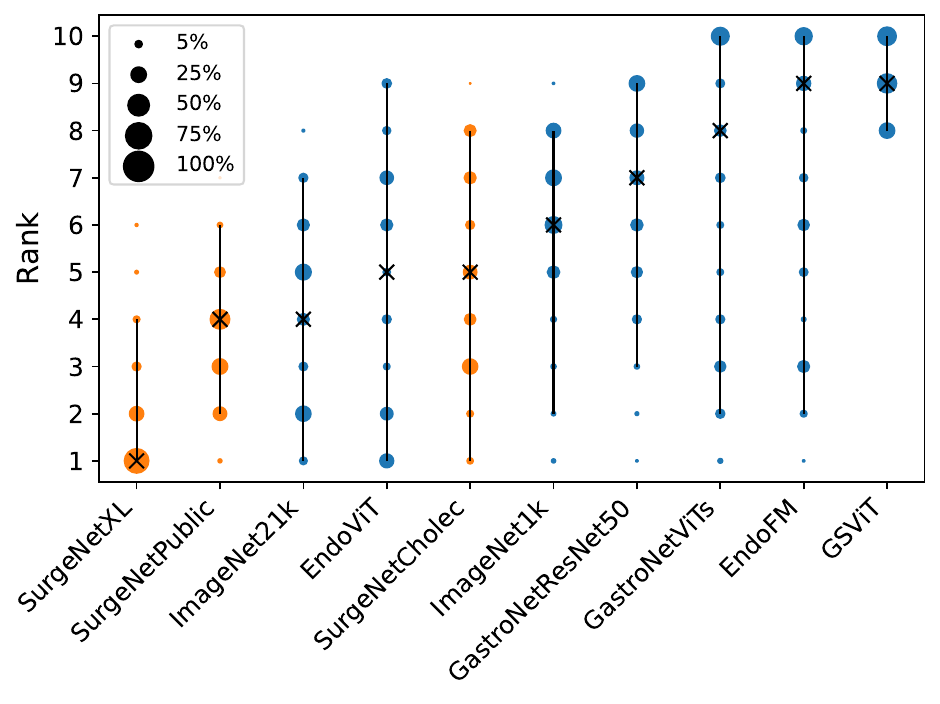}
     \caption{Ranking stability across all datasets and metrics. The size of each blob is proportional to the relative frequency with which a model architecture achieves a specific rank. The SurgeNetXL model (and its variations) are color-coded in orange. The median rank for each architecture, rounded to the nearest integer, is indicated by a black cross, while 95\% bootstrap intervals (spanning the 2.5th to 97.5th percentiles of the bootstrap distribution) are shown as black vertical lines. Models are ordered from left to right, with the best-performing model on the left and the worst on the right, based on the mean rank score across bootstrap samples.}
     \label{fig:ranking_all}
\end{figure}

Fig.~\ref{fig:ranking_all} presents the overall ranking of all models across all datasets. SurgeNetXL outperforms other models, followed by SurgeNetPublic. Among publicly available foundation models, EndoViT performs best, while GSViT ranks lowest.

\begin{figure*}[t]
     \centering
     \includegraphics[width=\textwidth]{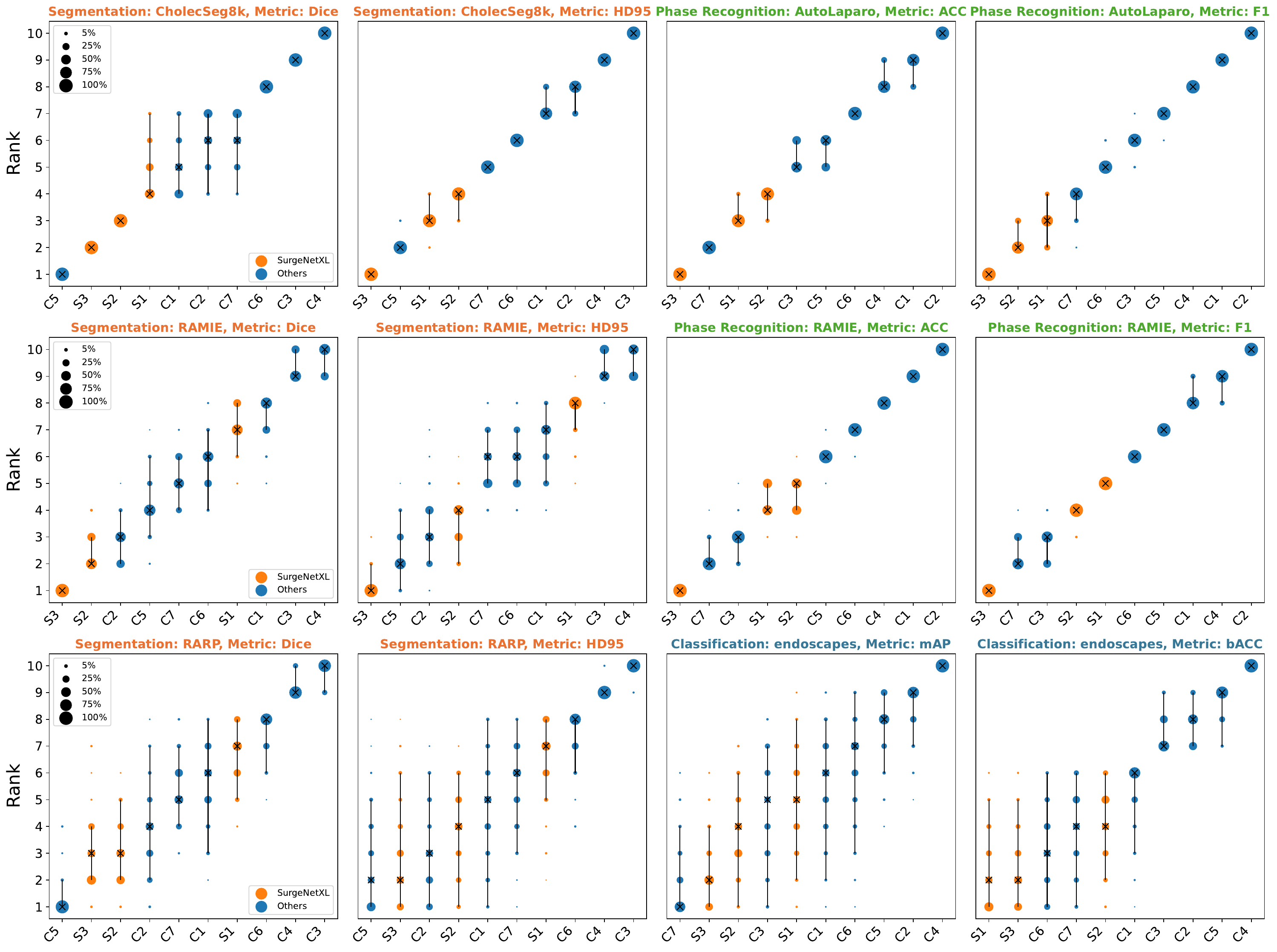}
     \caption{Ranking stability across each dataset and metric. The size of each blob is proportional to the relative frequency with which a model architecture achieves a specific rank. The SurgeNetXL model (and its variations) are color-coded in orange. The median rank for each architecture is indicated by a black cross, while 95\% bootstrap intervals (spanning the 2.5th to 97.5th percentiles of the bootstrap distribution) are shown as black vertical lines. Models are ordered from left to right, with the best-performing model on the left and the worst on the right, based on the mean rank score across bootstrap samples. The plot titles indicate the datasets used: semantic segmentation (orange), phase recognition (green), and CVS classification (blue). The model code names displayed on the x-axis are listed in Table~\ref{tab:sota_results}.}
     \label{fig:main}
\end{figure*}

To provide deeper insights, task- and metric-specific ranking stability plots are shown in Fig.~\ref{fig:main}. These results highlight the robustness of SurgeNetXL~(S3), which consistently ranks in the top 2 across all tasks and metrics. Conversely, EndoViT~(C6), despite excelling on the CholecSeg8k test set for Dice (Fig.\ref{fig:main}, first row, first column), exhibits lower consistency, ranking 8th and 9th for mAP and bACC, respectively, on the EndoScapes dataset (Fig.\ref{fig:main}, last row, blue titles).

While SurgeNetPublic achieves the second-best overall ranking, it displays greater variability in task-specific performance. For example, on the RAMIE dataset phase recognition task, it ranks 5th and 4th for ACC and F1 scores, respectively (second row, green titles). This variability may stem from the absence of esophagectomy data in the publicly available training datasets. A similar trend is observed with SurgeNetCholec, which delivers competitive performance on the CholecSeg8k and EndoScapes datasets but underperforms on the RAMIE and RARP segmentation datasets (second and third row, orange titles).

\subsection{Ablation experiments}
\label{sec:ablations}
As described in Section~\ref{sec:experimental_setup}, all ablation studies are performed using semantic segmentation data sets. These experiments aim to provide deeper insights into the critical factors that drive the development of a high-performing surgical foundation model. For these studies, SurgeNet is used as the base dataset instead of SurgeNetXL for two key reasons. First, using SurgeNet makes the SSL experiments more computationally feasible. Second, SurgeNet exclusively comprises verified datasets, with each frame precisely linked to its corresponding surgical procedure, a detail that is crucial for certain ablation experiments.

\subsubsection{Under-represented classes} 
\begin{figure*}[h]
     \centering
     \includegraphics[width=\textwidth]{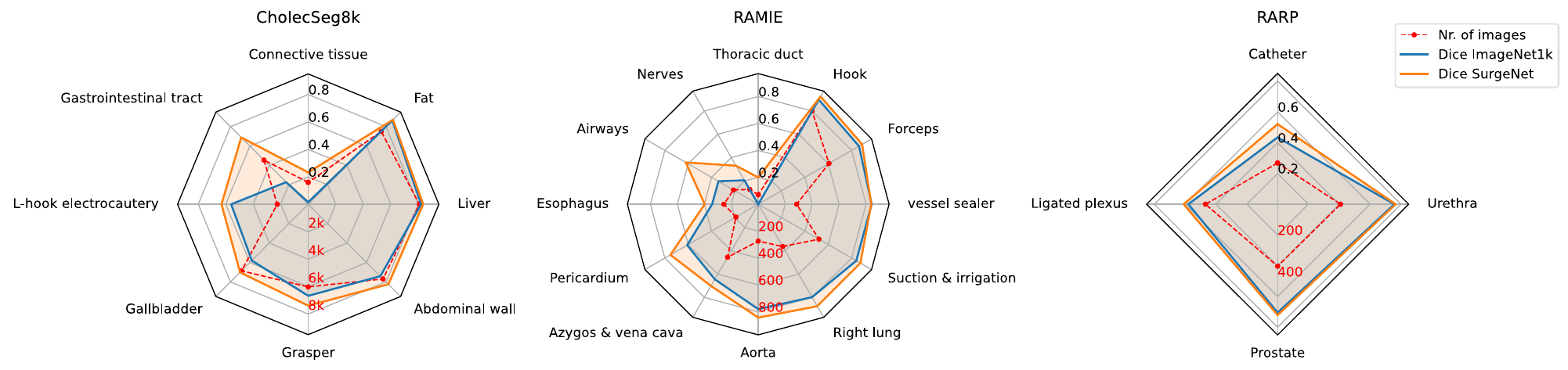}
     \caption{Mean Dice scores for each class on the CholecSeg8k, RAMIE, and RARP datasets are displayed from left to right, respectively. CAFormer, pretrained on ImageNet1k, is represented in orange, while SurgeNet is shown in blue. The number of images per class in each downstream dataset is indicated by the dotted red line.}
     \label{fig:results_per_class}
\end{figure*}

The radar plots in Fig.~\ref{fig:results_per_class} depict mean Dice scores for each class across the three segmentation datasets. The smallest performance gap between SurgeNet and ImageNet1k is observed for non-anatomy classes, such as surgical tools, and larger anatomical structures like fat, liver, lung, and urethra. In contrast, SurgeNet demonstrates the most significant performance gains in classes that are under-represented in terms of frequency and structure size, including nerves, the thoracic duct, connective tissue, and the catheter. This trend is further confirmed by the boxplots in Fig.~\ref{fig:boxplots_per_class}~(supplementary materials). This highlights the benefits of in-domain SSL, especially for classes that are difficult to annotate accurately but are essential for enhancing anatomy recognition systems. Visual examples are provided in the supplementary materials in Fig.~\ref{fig:visual_results}.

\subsubsection{Pretraining dataset composition}
\begin{figure*}[h]
     \centering
     \includegraphics[width=0.86\textwidth]{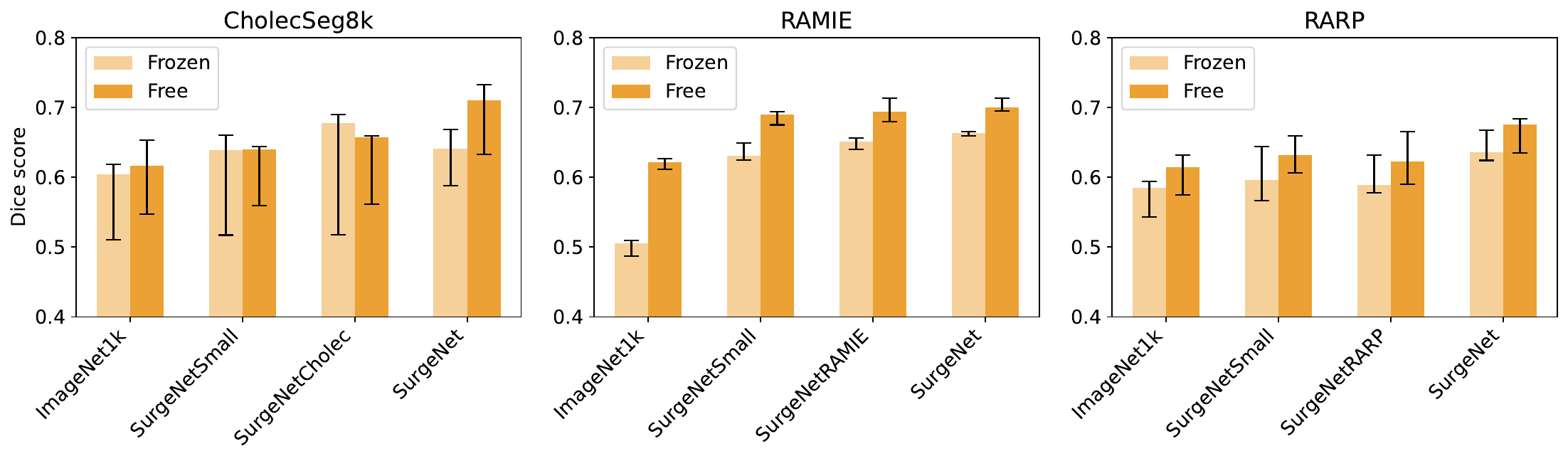}
     \caption{Dice scores on the  CholecSeg8k, RAMIE, and RARP datasets for ImageNet1k and various SurgeNet variants. Frozen indicates that the model is fine-tuned with a frozen encoder, while free signifies the use of an unfrozen encoder. Results are presented as medians, with error bars indicating the minimum and maximum values from the cross-validation folds.}
     \label{fig:pretraining_datasets}
\end{figure*}

Fig.~\ref{fig:pretraining_datasets} presents the results of SSL pretraining on SurgeNet and its procedure-specific datasets across the three downstream tasks. The encoder is evaluated with both frozen and trainable weights during downstream training. For laparoscopic cholecystectomy, RAMIE, and RARP, procedure-specific pretraining results in segmentation improvements of 6.7\%, 11.6\%, and 1.5\%, respectively, compared to ImageNet1k initialization with trainable encoder weights. Against ImageNet1k with frozen weights, improvements are 12.3\%, 28.9\%, and 0.7\%. These findings are in line with the results of \cite{alapatt2023}, who also observe substantial improvements using procedure-specific pretraining datasets.

For CholecSeg8k and RAMIE, our results indicate that training on procedure-specific datasets provides superior performance compared to SurgeNetSmall, despite their comparable sizes. Moreover, this study indicates that incorporating extra, more heterogeneous data during pretraining further enhances segmentation performance compared to procedure-specific training only. More specifically, training on SurgeNet results in a further improvement of 8.1\%, 0.9\%, and 8.3\% for laparoscopic cholecystectomy, RAMIE, and RARP, respectively, when encoder weights are trainable, albeit at the expense of a longer training time. Furthermore, the disparity between ImageNet1k and SurgeNet initialization is most pronounced on the CholecSeg8k and RAMIE datasets, which have the largest collection of difficult classes, such as small anatomical structures. This highlights SurgeNet's effectiveness on more challenging tasks. However, even on datasets with relatively large amounts of labeled data, such as CholecSeg8k, the segmentation model still benefits substantially from SurgeNet pretraining. 

Table~\ref{tab:self-supervised finetuning} shows the impact of procedure-specific fine-tuning of the SurgeNet model during pretraining. By self-supervised fine-tuning, we refer to initially pretraining the model on SurgeNet, followed by further pretraining on a procedure-specific dataset. Obtained results indicate that self-supervised fine-tuning does not yield any significant positive effect on performance.

\begin{table}[h]
    \scriptsize
    \setlength{\tabcolsep}{4pt}
    \centering
    \begin{tabular}{l | c c c c c c  }
        \hline
        \toprule
        Pretraining Dataset & \multicolumn{2}{c}{CholecSeg8k} & \multicolumn{2}{c}{RAMIE} & \multicolumn{2}{c}{RARP} \\
                            & Median & [\%] & Median & [\%] & Median & [\%] \\ 
        \midrule 
        SurgeNet                  & 0.710 & -         & 0.700 & -       & 0.675 & -\\
        SurgeNet + SS fine-tuning & 0.709 & -0.1\%  & 0.704 & 0.6\% & 0.677 & 0.3\% \\
        \bottomrule
    \end{tabular}
    \caption{Median Dice scores and percentage-wise differences of pretraining on SurgeNet vs. adding self-supervised (SS) fine-tuning.}
    \label{tab:self-supervised finetuning}
\end{table}

\subsubsection{Architectures}
\begin{figure*}[h!]
     \centering
     \includegraphics[width=0.86\textwidth]{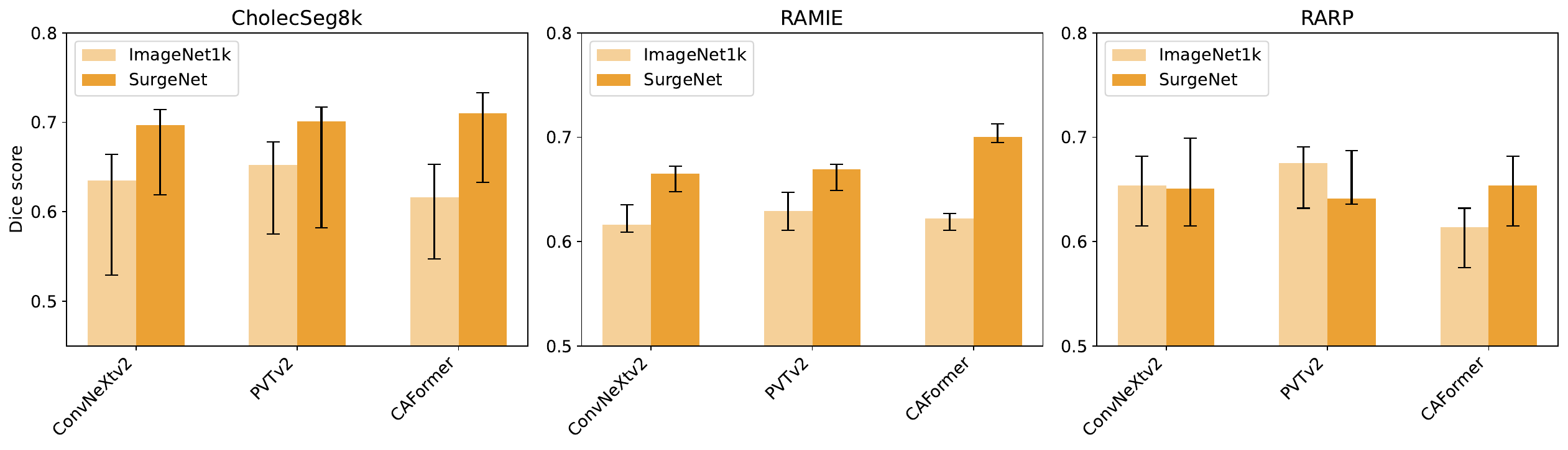}
     \caption{Dice scores on the CholecSeg8k, RAMIE, and RARP datasets for various encoders pretrained on ImageNet1k and SurgeNet. Results are presented as medians, with error bars indicating the minimum and maximum values from the cross-validation folds.}
     \label{fig:different_encoders}
\end{figure*}

Fig.~\ref{fig:different_encoders} presents the results of evaluating the impact of SurgeNet pretraining on different encoder architectures. Across the three datasets, we observe mean performance improvements of 7.1\%, 1.9\%, and 0.2\% for the three encoders compared to ImageNet1k pretraining. The CAFormer architecture, chosen as the basis for the SurgeNetXL foundation model, shows the most significant gains from pretraining. However, both the CNN-based ConvNeXtV2 and the vision transformer-based PVTv2 encoders also demonstrate substantial improvements, except on the RARP dataset. These results suggest that SurgeNet pretraining provides broad benefits, enhancing performance across a wide range of model architectures.

\subsubsection{Pretraining time}
\begin{figure*}[h]
     \centering
     \includegraphics[width=\textwidth]{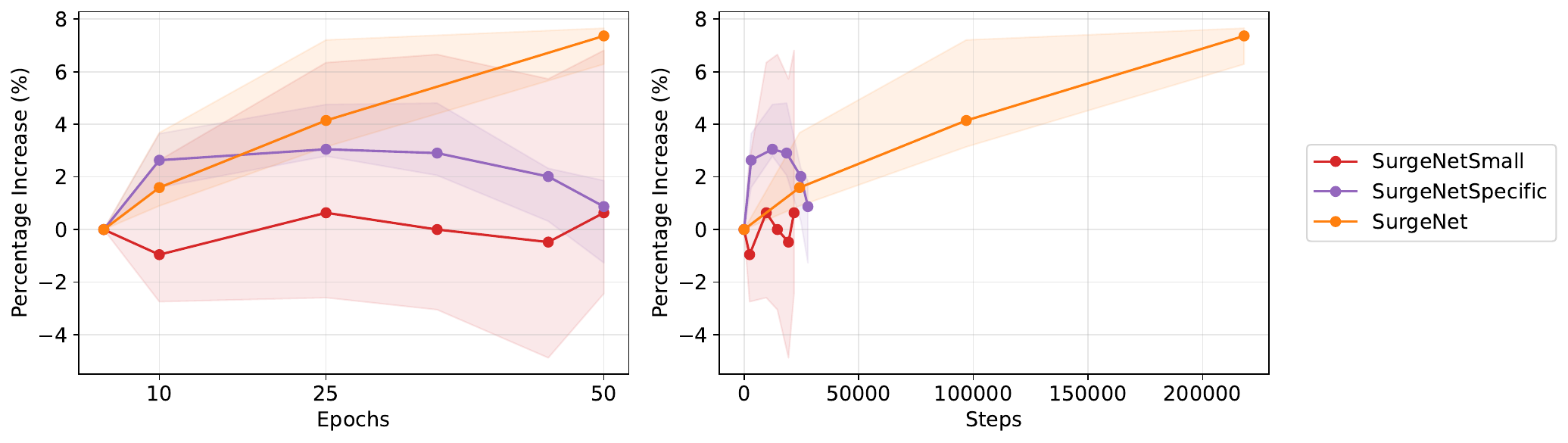}
     \caption{Left: Percentage-wise Dice score increase averaged over all segmentation datasets during pretraining with DINO, relative to epoch 5. Right: The same plot, but with the x-axis representing pretraining steps. The SurgeNet line extends further because the same number of epochs is applied to a larger dataset, resulting in more steps. SurgeNetSpecific denotes the average Dice score increase for the procedure-specific pretraining datasets (SurgeNetCholec, SurgeNetRAMIE, and SurgeNetRARP). Results are presented as medians, with shaded areas indicating the minimum and maximum values from the cross-validation folds.}
     \label{fig:dino_train_time}
\end{figure*}

Fig.~\ref{fig:dino_train_time} illustrates the relative performance gains across checkpoints during pretraining. The results show that pretraining on SurgeNet continues to yield improvements even beyond 50 epochs, highlighting its capacity for sustained learning. In contrast, pretraining on procedure-specific datasets provides notable performance gains during the initial epochs but decreases after 25 epochs, suggesting limited diversity in these datasets. The SurgeNetSmall variant shows no significant improvement beyond epoch 5, highlighting the importance of dataset size in SSL.

\subsubsection{Downstream data size}
\begin{figure*}[h!]
     \centering
     \includegraphics[width=\textwidth]{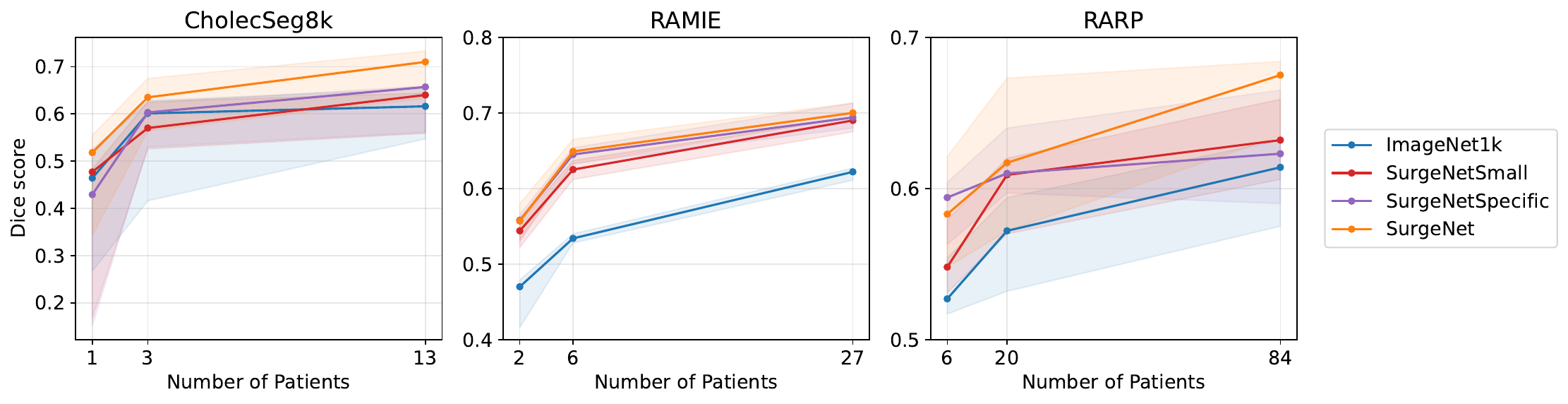}
     \caption{Dice scores versus the number of training patients for the CholecSeg8k, RAMIE, and RARP datasets. SurgeNetSpecific refers to SurgeNetCholec, SurgeNetRAMIE, and SurgeNetRARP, from left to right, respectively. Results are presented as medians, with shaded areas indicating the minimum and maximum values from the cross-validation folds.}
     \label{fig:downstream_data}
\end{figure*}

Fig.~\ref{fig:downstream_data} illustrates the impact of varying downstream training data size on performance. The results demonstrate that the SurgeNet variants mostly outperform ImageNet1k, especially on the RAMIE and RARP datasets. Using SurgeNet pretraining on RAMIE with 6 patients outperforms ImageNet1k pretraining with 27 patients. On RARP the same observation holds, with only 20 patients compared to 84. This indicates that SurgeNet can reduce the burden of annotating training data on complex datasets. For CholecSeg8k, the differences in the Dice score are relatively small and the impact of self-supervised learning is less noticeable. However, SurgeNet still achieves better performance than ImageNet1k.

\subsubsection{Unsupervised clustering}
\begin{figure*}[h] 
     \centering
     \includegraphics[width=0.9\textwidth]{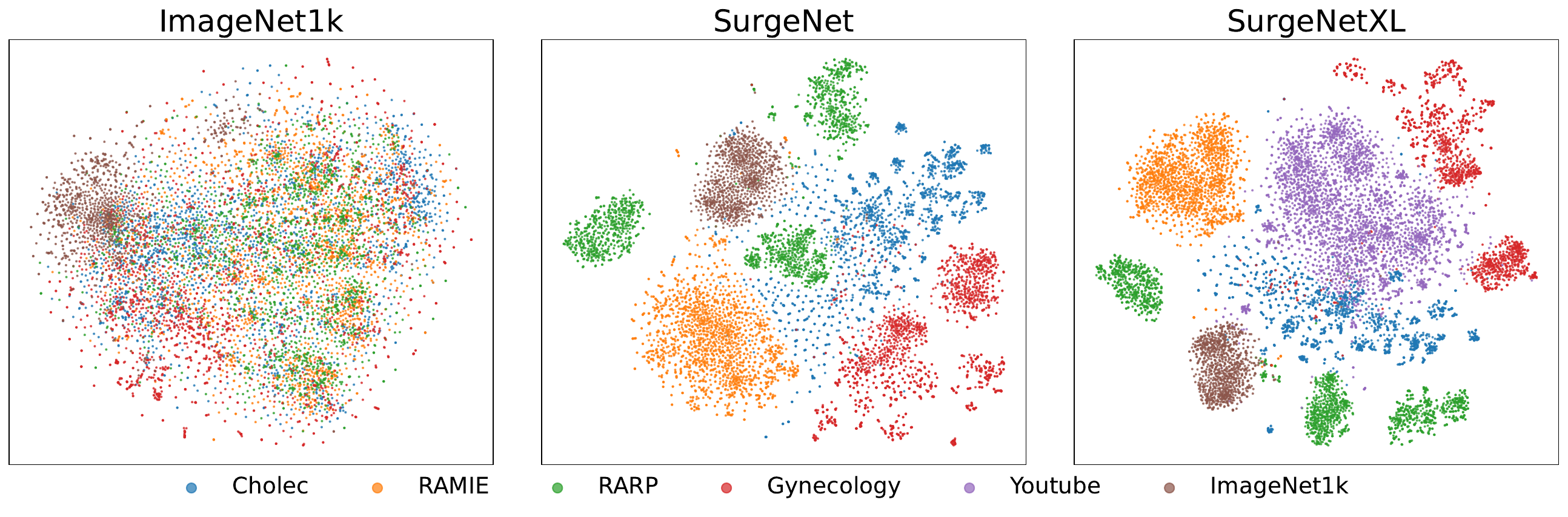}
     \caption{t-SNE visualization of CAFormer using different weight initializations, from left to right: ImageNet1k, SurgeNet, and SurgeNetXL. Each dataset source is represented by a distinct color.}
     \label{fig:tSNE}
\end{figure*}

Fig.~\ref{fig:tSNE} illustrates the 2D t-SNE visualization of the CAFormer encoder with different pretrained weights, highlighting the model's ability to represent video frames from various surgical procedures.

The left panel shows the visualization with ImageNet1k-initialized weights, where no clear clusters are observed, indicating the encoder's limited ability to distinguish between frames from different surgical contexts. In contrast, the middle panel, which utilizes SurgeNet-initialized weights, demonstrates well-defined clusters, with images from the same procedure grouped closely together. This clustering, achieved without any supervision during pretraining, highlights that the encoder trained on the SurgeNet dataset has learned procedure-specific representations. Such meaningful embeddings suggest a strong potential for generalization to unseen data, thereby enhancing the model's performance in downstream tasks. The results  emphasize the importance of pretraining on a diverse and comprehensive dataset like SurgeNet, which equips the model with a robust understanding of surgical contexts.

The right panel provides further insights through the t-SNE visualization using the SurgeNetXL encoder, which incorporates additional data from the surgical YouTube dataset. While images from established datasets—typically high-quality data collected at single centers or at a few expert institutions—form clear and compact clusters, the YouTube dataset does not exhibit similarly distinct clustering. This reflects the diverse and heterogeneous nature of the YouTube dataset, encompassing a broader range of surgical contexts. 

\section{Discussion}
\label{sec:discussion}
 This study explores the impact of large-scale pretraining for surgical computer vision using SSL on the largest reported surgical dataset to date. Our extensive benchmark demonstrates the robustness and generalizability of SurgeNetXL, which achieves top-2 performance across all metrics on six downstream datasets, including three different tasks and four surgical procedures. Unlike other foundation models that exhibit variability in performance depending on the dataset, SurgeNetXL maintains consistent performance, underscoring its broad applicability. The inclusion of the Surgical YouTube dataset as an extension to the existing dataset repertoire in particular emerges as a key contribution to its success. Through ablation experiments, we further examine the impact of pretraining time and dataset composition, demonstrating that SSL is effective across various model architectures.

\textit{Benchmark: }
The extensive benchmarking, against five different surgical foundation models and four other variations of SurgeNetXL, highlights the advantages of SurgeNetXL in multiple surgical computer vision tasks, including semantic segmentation, phase recognition, and CVS classification. Compared to the best-performing surgical foundation models, mean improvements of 2.4, 9.0, and 12.6 percent are found for semantic segmentation, phase recognition, and CVS classification, respectively, when averaged over all metrics and datasets. Additionally, improvements of 14.4, 4.0, and 1.6 percent are found compared against the best-performing ImageNet variation.  This indicates that the model is well-suited to varied and complex downstream tasks. In contrast, other foundation models like EndoViT~\citep{Batic2024}, show competitive performance on specific datasets (e.g., CholecSeg8k), but score approximately 10\% lower on other datasets compared to SurgeNetXL. This consistency across all tasks emphasizes the critical role of pretraining on diverse and comprehensive datasets in developing an effective surgical foundation model.

\textit{Surgical YouTube dataset: }
The Surgical YouTube dataset, introduced and open-sourced as part of this study, is manually curated to ensure its validity. It provides substantial improvements as an extension to existing datasets. The YouTube dataset is more diverse in terms of the procedures included, the quality, and the number of videos compared to the existing datasets. Even combining all publicly available datasets collected for this study does not reach the same number of procedures and videos. While the size of the YouTube dataset is comparable to other large-scale natural image datasets like ImageNet, its diversity and variation remain limited in comparison, reflecting the inherent challenges and specificity of surgical computer vision data. Performance compared to SurgeNetPublic increases by 7.3\%, 3.7\%, and 3.6\% averaged over datasets and metrics for semantic segmentation, phase recognition, and CVS classification, respectively. These findings highlight the impact of the YouTube dataset on the field of SSL for surgical computer vision.

\textit{Ablations: }
Further insights from ablation experiments highlight the critical influence of dataset composition during pretraining. Pretraining on SurgeNet and its variations consistently outperforms ImageNet1k, particularly in scenarios involving under-represented anatomy classes or limited labeled data. The advantages are most pronounced in datasets like RAMIE and CholecSeg8k, which contain a large number of difficult classes. In particular, pretraining on procedure-specific data alone yields greater benefits compared to SurgeNetSmall, which aligns with the findings from previous research by \cite{alapatt2023}. However, incorporating frames from diverse procedures and pretraining on a more comprehensive dataset like SurgeNet consistently enhances performance, suggesting the value of broader dataset diversity.

The above mentioned improvements extend across diverse encoder architectures, including ConvNeXtV2, PVTv2, and CAFormer, showcasing the generalizable benefits of SurgeNet pretraining across different model types. These architectures are specifically chosen because they represent SOTA approaches within their respective categories: ConvNeXtV2 for CNNs, PVTv2 for transformer-based models, and CAFormer as a hybrid architecture. This selection allows us to comprehensively evaluate the impact of pretraining on diverse architectural paradigms. Among these, the CAFormer architecture, adopted as the foundation for SurgeNetXL, achieves the most significant performance gains. Please note that we opted not to use the largest variants of these architectures, as doing so would substantially increase the computational cost of the experiments and severely reduce batch sizes, a critical factor for the success of SSL methods.

Additionally, Fig.\ref{fig:dino_train_time} indicates that pretraining on SurgeNet does not converge after 50~epochs, suggesting that further improvements could be achieved with extended training. In contrast, smaller datasets, such as SurgeNetSmall and procedure-specific variations, appear to reach their optimal performance within the 50-epoch timeframe.

The t-SNE visualizations provide further validation of these findings. Models pretrained on SurgeNet develop meaningful and procedure-specific embeddings that enhance generalization. This is evidenced by the creation of distinct, procedure-specific clusters even in the absence of supervised information. These results highlight the capacity of SurgeNet-pretrained models to learn robust and transferable representations, particularly valuable in complex medical imaging tasks.

\textit{Computational resources: }
The pretraining experiments presented in this study require over 6500~GPU hours, excluding the additional computational resources needed for downstream evaluations. This highlights the substantial computational demands of such experiments. Consequently, by making these models publicly available, we enable other researchers to build on this work.

While prior research on surgical foundation models has largely compared ImageNet-initialized weights across various downstream datasets, it has not yet delved deeply into other factors influencing SSL for surgery. This is likely due to the vastly higher computational demands of pretraining experiments compared to downstream evaluations. Notably, \cite{RAMESH2023} laid foundational work for exploring SSL for surgical deep learning models. However, their experiments were conducted on a significantly smaller scale, with a pretraining dataset over 20~times smaller than SurgeNetXL presented in this study. This work builds upon their contributions by addressing the challenges and opportunities of large-scale pretraining in the surgical domain.

\textit{Limitations: }
A key limitation of this study lies in the lack of tailored optimization for the training procedures of the evaluated models. To maintain consistency and isolate the impact of pretrained weights, we follow a standardized approach, utilizing a limited set of data augmentation techniques and predefined learning rates with widely used scheduling strategies. While this choice minimizes confounding variables, it may constrain the models' potential to achieve optimal performance. Future work could explore more adaptive training strategies, customized augmentation pipelines, and dynamic learning rate adjustments to fully utilize the capabilities of each model.

Another limitation lies in the architectural design of downstream applications. Since not all SOTA models provided frameworks suitable for every task in our study, we adjusted these models accordingly. To ensure a fair comparison, we apply the same architectural choices to these models as we do to SurgeNetXL. Specifically, we use the FPN decoder for semantic segmentation and the MS-TCN for phase recognition. However, alternative decoder architectures may be more advantageous for these models.

\textit{Future directions: }
Building on recent advances in general-purpose visual pretraining, such as DINOv2~\citep{oquab2024dinov2learningrobustvisual}, future research could investigate its applicability in surgical computer vision. Although this study does not explore alternative pretraining methods such as SimCLR, MoCoV2, or MAE—given previous findings that these approaches yield similar benefits \citep{BOERS2024, RAMESH2023}—DINOv2 represents a significant leap in the natural image domain, offering improved robustness and generalizability. Evaluating its impact in the surgical domain could uncover new opportunities to advance pretraining strategies tailored to medical imaging tasks.

Furthermore, the potential of video-based SSL remains untouched in this study. Temporal dynamics and motion cues are critical in surgical workflows, but are often overlooked in frame-by-frame approaches. Recent innovations, such as V-JEPA~\citep{bardes2024revisiting}, effectively capture temporal relationships in general computer vision tasks. Extending such methods to surgical applications could lead to substantial improvements in tasks reliant on motion understanding, such as phase recognition and tool tracking. However, it is important to note that most current frameworks rely on frame-based approaches. Transitioning to video-based SSL may require overcoming practical challenges, including higher computational demands and limited availability of large-scale annotated surgical video datasets. These constraints may temper the immediate impact of video-based SSL but should not diminish its potential for long-term advancements in the field.

\section{Conclusion}
\label{sec:conclusion}
In conclusion, this study demonstrates the potential of large-scale, diverse pretraining to advance surgical computer vision. While foundation models have shown promise in medical computer vision, their application in the surgical domain has remained underexplored, with prior models relying on significantly smaller pretraining datasets compared to those used in natural image tasks. Furthermore, no comprehensive benchmarking has been conducted since these models were introduced in quick succession. Addressing these gaps, SurgeNetXL, a novel surgical foundation model, achieves top-2 performance across all metrics on six downstream datasets spanning three tasks and four procedures.

A critical factor in SurgeNetXL's success is the inclusion of the Surgical YouTube dataset, a major contribution of this work, which comprises over 2 million frames from more than 3,000 surgical videos. This dataset open-sourced alongside all models as part of this work, not only enhances diversity but also represents the largest surgical dataset to date, underscoring the importance of scale and variety in pretraining datasets for improving model performance. Furthermore, this study highlights the robustness and generalizability of SurgeNetXL, demonstrating its effectiveness across diverse tasks and model architectures. Together, these findings mark a significant step forward in leveraging foundation models to advance the field of surgical computer vision.

\section*{Acknowledgments}
This research is funded by Stichting Hanarth Fonds, study number: 2022-13. It is part of the INTRA-SURGE (INTelligent computeR-Aided Surgical gUidance for Robot-assisted surGEry) project aimed at advancing the future of surgery. Furthermore, we thank SURF (\url{www.surf.nl}) for the support in using the National Supercomputer Snellius.

\bibliographystyle{model2-names.bst}\biboptions{authoryear}
\bibliography{refs}

\section{Supplementary Materials}
\subsection{Pretraining details}
Table~\ref{tab: dino details} outlines the hyperparameters employed during DINO pretraining, which are closely aligned with the originally proposed configurations by \cite{caron2021emerging}. The data augmentation techniques utilized in the process include random horizontal flipping, color jittering, grayscale transformation, and Gaussian blurring.

\begin{table}[h!]
\centering
    \caption{Training details of DINO pertaining on SurgeNetXL and its variations.}
        \begin{tabular}{ l | c }
             \toprule
             {Hyperparameter}  & { Value} \\            
             \midrule
             Input size             & 224 \\
             Optimizer              & SGD \\
             Gradient clip          & 3.0 \\
             LR decay schedule      & cosine schedule \\
             Train epochs           & 50 \\
             Train batch size       & 544 \\
             learning rate          & $5\times10^{-4}$ \\
             Warming-up epochs      & 10 \\
             Weight decay           & 0.04 \\
             Mixed precision        & yes \\
             \bottomrule
             
        \end{tabular}
    \label{tab: dino details}
\end{table}

\subsection{SOTA model fine-tuning details}
When fine-tuning the SOTA models, we maintain all training parameters identical to those used for SurgeNet to ensure consistency. However, minor model adjustments are made to align with specific downstream requirements. For semantic segmentation, GastroNet, Endo-FM, and EndoViT employ Mobile DeepLabv3+~\citep{sandler2018mobilenetv2}, TransUNet~\citep{chen2021transunet}, a Dense Prediction Transformer~\citep{ranftl2021vision} as decoder, respectively, adhering to their original methodologies. Since GSViT was not originally designed for semantic segmentation, we incorporate the same FPN decoder used in the SurgeNet models. For phase recognition and CVS classification tasks, the decoders remain identical to those utilized in the SurgeNet models.

\subsection{Results per class}
\begin{figure*}[h]
     \centering
     \includegraphics[width=0.86\textwidth]{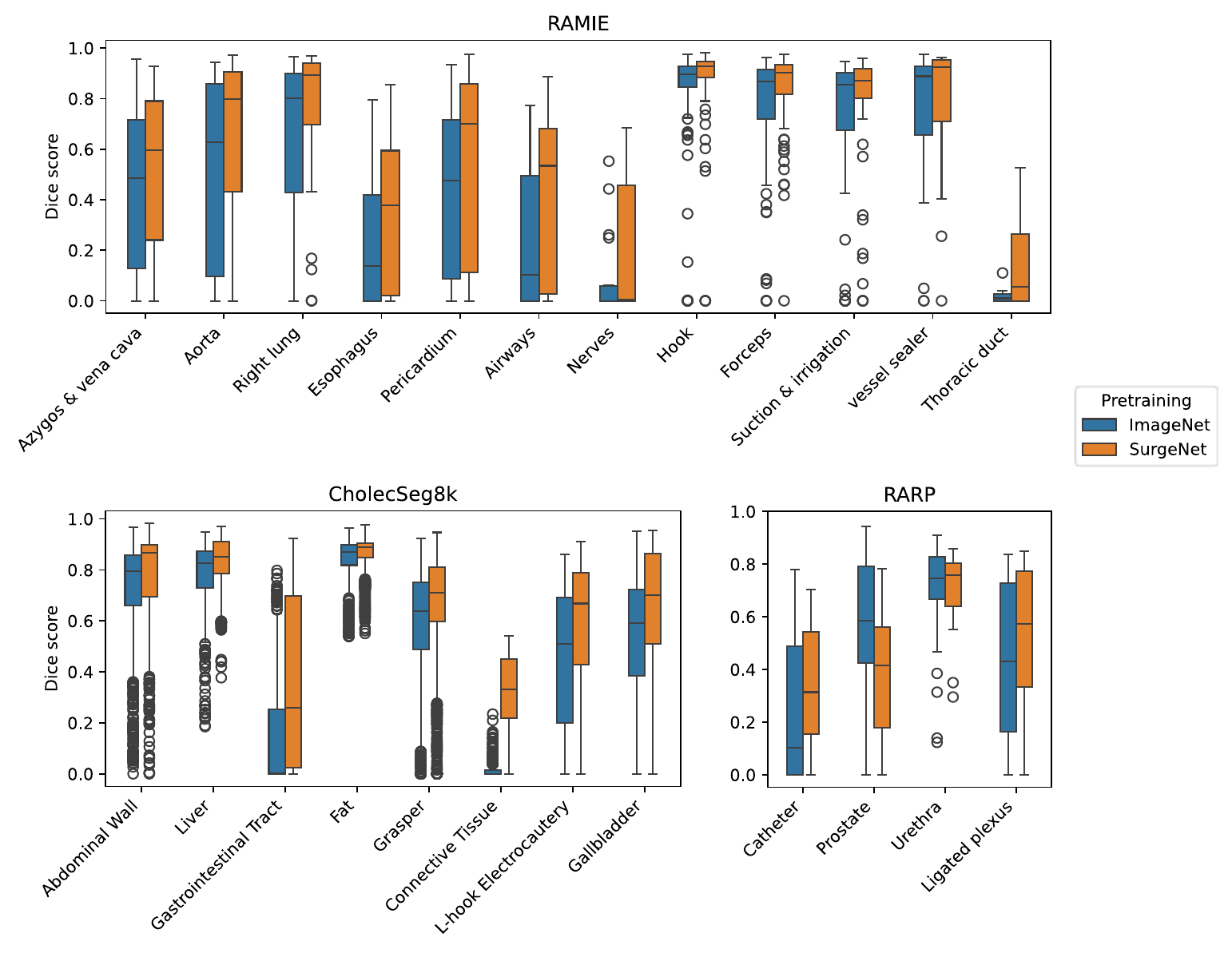}
     \caption{Dice scores per class for SurgeNet and ImageNet initialization across all three semantic segmentation datasets.}
     \label{fig:boxplots_per_class}
\end{figure*}

Figure~\ref{fig:boxplots_per_class} presents boxplots for each class across the three semantic segmentation datasets. These visualizations show that certain classes derive greater benefits from SurgeNet pretraining compared to others. Notably, in the RAMIE dataset, challenging classes such as the nerves and thoracic duct exhibit significant improvements in median performance. Similarly, the connective tissue class in the CholecSeg8k dataset demonstrates a marked increase in performance. However, the boxplots also highlight substantial variation in Dice scores across the test sets.  

\subsection{Visual results}
\begin{figure*}[h!]
     \centering
     \includegraphics[width=\textwidth]{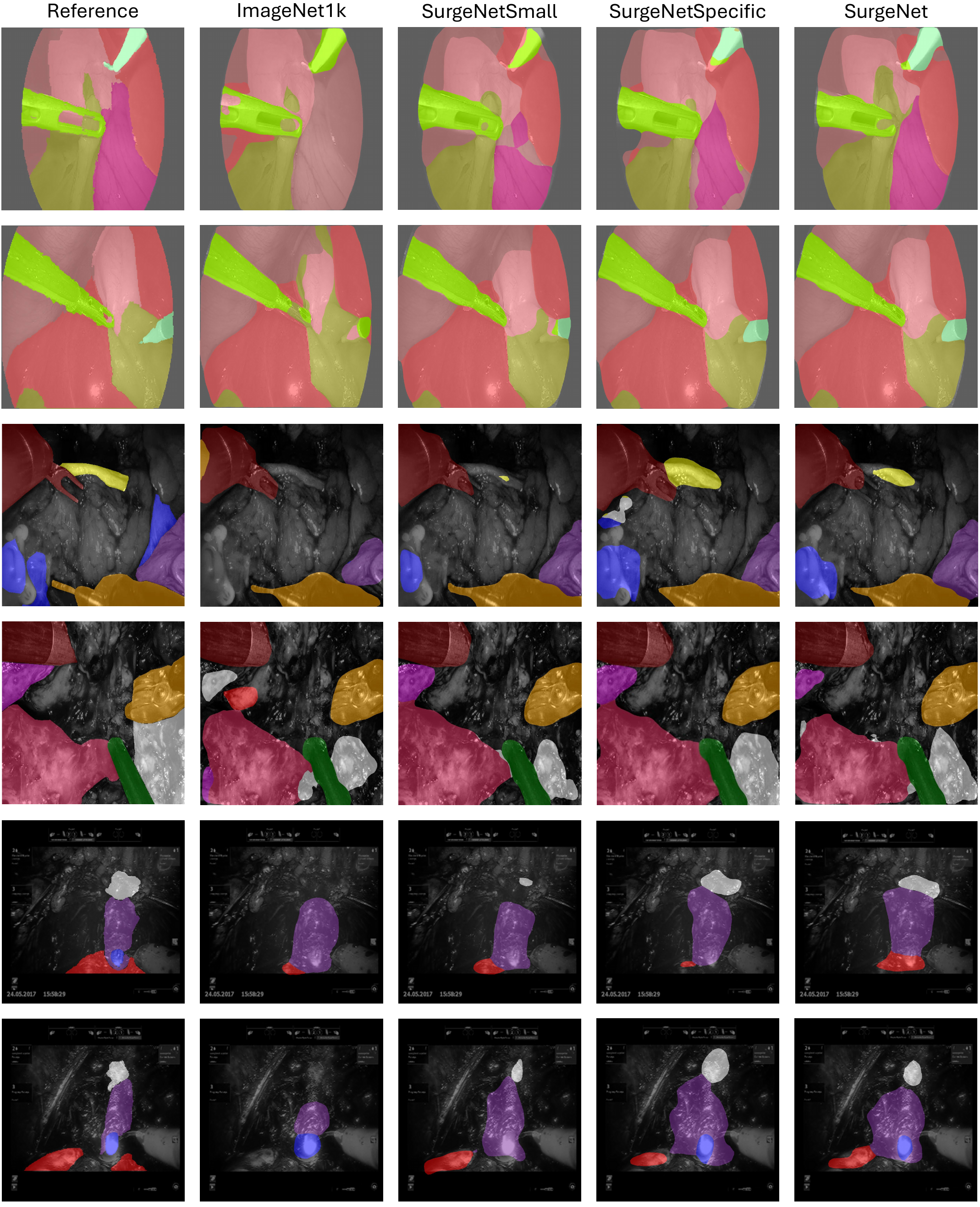}
     \caption{Visual examples including the three downstream datasets for semantic segmentation. The first two rows display samples from the CholecSeg8k test set, the subsequent two rows show samples from the RAMIE test set, and the final two rows present samples from the RARP test set.}
     \label{fig:visual_results}
\end{figure*}

Fig.~\ref{fig:visual_results} illustrates the visual results for the three downstream semantic segmentation datasets. Comparing models trained on ImageNet1k and SurgeNet variants, it is evident that SurgeNet better captures challenging structures for segmentation. For example, in the first row, the gastrointestinal tract, shown in purple, is identified by SurgeNet but not by the ImageNet1k model. Similarly, the nerve, depicted in yellow in the third row, and the ligated plexus, shown in white in the fifth and sixth rows, are accurately segmented only by SurgeNet variants. These visual results further demonstrate that SurgeNet achieves superior performance in segmenting challenging classes, which is in line with our quantitative findings.

\end{document}